\documentclass{article}
\PassOptionsToPackage{numbers, compress}{natbib}

    \usepackage[preprint]{neurips_2023}

\usepackage[utf8]{inputenc} 
\usepackage[T1]{fontenc}    
\usepackage{hyperref}       
\usepackage{url}            
\usepackage{booktabs}       
\usepackage{amsfonts}       
\usepackage{nicefrac}       
\usepackage{microtype}      
\usepackage{xcolor}         
\usepackage{amsmath, amssymb} 
\usepackage{amsthm} 
\usepackage{tikz}
\usepackage{bm}
\usetikzlibrary{automata, positioning, arrows}
\usepackage{environ}
\usepackage{subcaption}
\usepackage{multirow}
\usepackage{colortbl}
\usepackage{makecell}
\usepackage{float}
\usepackage{wrapfig}
\usepackage{graphicx} 

\theoremstyle{plain}
\newtheorem{theorem}{Theorem}[section]
\newtheorem{proposition}[theorem]{Proposition}

\theoremstyle{definition}
\newtheorem{definition}[theorem]{Definition}
\newtheorem{lemma}[theorem]{Lemma}

\newtheorem{assumption}[theorem]{Assumption}
\theoremstyle{remark}

\newcommand{\lbbr}{\{\!\!\{}
\newcommand{\rbbr}{\}\!\!\}}

\title{Fine-Grained Expressive Power of Weisfeiler-Leman: A Homomorphism Counting Perspective}
\author{
    Junru Zhou$^1$\\
    \texttt{zml72062@stu.pku.edu.cn}\And Muhan Zhang$^1$\\
    \texttt{muhan@pku.edu.cn}
}
\date{}

\begin{document}

\maketitle

\begin{abstract}
    The ability of graph neural networks (GNNs) to count homomorphisms has recently been proposed as a practical and fine-grained measure of their expressive power. Although several existing works have investigated the homomorphism counting power of certain GNN families, a simple and unified framework for analyzing the problem is absent. In this paper, we first propose \emph{generalized folklore Weisfeiler-Leman (GFWL)} algorithms as a flexible design basis for expressive GNNs, and then provide a theoretical framework to algorithmically determine the homomorphism counting power of an arbitrary class of GNN within the GFWL design space. As the considered design space is large enough to accommodate almost all known powerful GNNs, our result greatly extends all existing works, and may find its application in the automation of GNN model design.
\end{abstract}

\section{Introduction}\label{sect-intro}

Graph Neural Networks (GNNs) have achieved remarkable success in the field of graph representation learning. However, since no deterministic polynomial-time algorithm for graph isomorphism testing has yet been found~\citep{babai2016graph}, none of the popular GNN models are universal approximators of graph functions even with sufficient number of parameters, according to~\citet{chen2019equivalence}. This drawback of GNNs has been informally stated as the \textbf{limited expressive power} of GNNs. 

In particular,~\citet{xu2018powerful} has proved that the expressive power of Message Passing Neural Networks (MPNNs), a widely adopted class of GNNs, is upper-bounded by that of 1-dimensional Weisfeiler-Leman (WL) algorithm, or 1-WL~\citep{weisfeiler1968reduction}. Following~\citet{xu2018powerful}, numerous works have proposed more powerful GNNs based on higher-order WL algorithms as well as their variants~\citep{feng2022powerful, feng2023towards, huang2023boosting, morris2019weisfeiler, morris2020weisfeiler, qianordered2022, zhang2023complete, zhou2023relational, zhou2024distance}. Some of the works~\citep{morris2019weisfeiler, morris2020weisfeiler, qianordered2022, zhang2023complete, zhou2023relational} even systematically compare the expressive power of those powerful GNNs, and establish expressiveness hierarchies among them. However, these comparisons usually fail to quantify the expressiveness gap between different methods, and provide little insight into what tasks a certain class of GNNs tend to excel at.

Recently, there are several works~\citep{jin2024homomorphism, zhang2024weisfeilerlehman} trying to characterize the expressive power of GNNs by considering their ability to perform \textbf{homomorphism counting}, namely, the power to determine the number of homomorphisms from a small query graph to the graph we are interested in. Earlier on, another line of works~\citep{chen2020can, huang2023boosting, zhou2024distance} has proposed to study GNN expressiveness by considering their \textbf{substructure counting} power, or the power to determine subgraph \emph{isomorphisms} instead of homomorphisms. Both approaches can provide fine-grained knowledge about the expressive power of GNNs. As an example, the expressiveness gap between two families of GNNs can be made manifest by examining the substructures whose homomorphism (or isomorphism) counts that one of them can compute while the other cannot. Despite the similarity between those two perspectives, it is argued that homomorphism counting power is a better metric for expressiveness, due to reasons such as that homomorphism counts are known~\citep{curticapean2017homomorphisms} to be a good basis upon which many other graph parameters (including subgraph isomorphism counts) can be computed, and that there is an exact equivalence~\citep{barcelo2021graph, dell2018lovasz} between the ability to count homomorphisms for some certain classes of query graphs and the well-known Weisfeiler-Leman expressiveness hierarchy, while such equivalence does not exist for substructure counting.

As many popular GNNs are designed upon (and easily shown to be as expressive as) variants of WL algorithms, we may now confine our attention to the following question: \textbf{what exactly query graphs whose homomorphisms can a specific variant of Weisfeiler-Leman algorithm count?} While the terminologies in the above question are to be made clear later, we point out that answering the question is not easy. Even for the well-studied $k$-dimensional folklore Weisfeiler-Leman algorithms ($k$-FWL)~\citep{cai1992optimal, grohe2015pebble, maron2019provably}, determining their exact homomorphism counting power relies on the (not so intuitive) concept of tree decompositions~\citep{dell2018lovasz}. Another work~\citep{zhang2024weisfeilerlehman} closely follows~\citep{dell2018lovasz}, and extends its result to three other variants of WL algorithms using the language of nested ear decompositions. Despite those efforts, no answer to the above question has been given in the most general setting.

\paragraph{Main contributions.} We now summarize our main contributions as following:
\begin{enumerate}
    \item We introduce the concept of \textbf{generalized folklore Weisfeiler-Leman (GFWL) algorithms}, which encompasses most existing variants of WL algorithms, including those proposed in the latest works such as~\citep{feng2023towards, zhou2024distance}.
    \item We derive the exact set of query graphs whose homomorphisms an instance of generalized folklore Weisfeiler-Leman algorithm can count. In other words, we find that for any GFWL algorithm $\mathcal{A}$, $\mathcal{A}$-indistinguishability is always equivalent to identity in homomorphism counts from an appropriate set of graphs. Our results greatly extend those in~\citep{dell2018lovasz, zhang2024weisfeilerlehman}. Our proof of Theorem \ref{main_theorem} relies on the relationship between GFWL algorithms and two types of pebble games, namely \textbf{Ehrenfeucht-Fra\"iss\'e games} and \textbf{Cops-Robber games}, which have been well-known tools to analyze $k$-FWL algorithms ever since~\citep{cai1992optimal, ehrenfeucht1961application, fraisse1954quelques, seymour1993graph}. Our results show that these tools can be generalized to give a unified characterization of a much larger family of algorithms.
    \item Our main result, Theorem \ref{main_theorem}, actually provides us with a \emph{meta-algorithm}\footnote{We call it a \emph{meta-algorithm} since it takes as input and evaluates another algorithm.} to automate the procedure of determining the homomorphism counting power of any GFWL algorithm. We implement the meta-algorithm and validate its correctness by comparing its output with previous theoretical results in~\citep{arvind2020weisfeiler, feng2023towards, huang2023boosting, neuen2023homomorphism, zhang2023complete, zhang2024weisfeilerlehman, zhou2024distance}. Our code is present at \href{https://github.com/zml72062/pebble\_game}{\texttt{https://github.com/zml72062/pebble\_game}}.
\end{enumerate}

\paragraph{Limitations.} Although our proposed GFWL framework accommodates a number of existing powerful WL variants, such as $k$-FWL, local $k$-FWL~\citep{zhang2024weisfeilerlehman} and $\delta$-DRFWL(2)~\citep{zhou2024distance}, it still leaves some well-known variants out, like $k$-WL~\citep{morris2019weisfeiler} or sparse $k$-WL~\citep{morris2020weisfeiler}. Further, our current work focuses solely on homomorphism counting power, and does not discuss its connection with substructure counting power (the latter being more intuitive). Yet existing works~\citep{neuen2023homomorphism, zhang2024weisfeilerlehman} have shown that \textbf{at least for some specific WL variants, being able to count subgraph isomorphisms from a query graph $\bm{H}$ is equivalent to being able to count homomorphisms from all homomorphism images of $\bm{H}$}, indicating a deep connection exists between homomorphism counting and substructure (isomorphism) counting. While we conjecture that the above connection holds for general GFWL algorithms, we have difficulty proving it. Due to the aforementioned limitations, this paper serves as a preliminary work, and directions for future works are discussed in Section~\ref{sec:future}.

\section{Notations and definitions}

\subsection{Basic notations}

For any $n\in\mathbb{N}^*$, we denote $[n]=\{1,2,\ldots, n\}$. Let $\mathcal{G}$ be the set of all simple, undirected graphs. For a graph $G\in\mathcal{G}$, we use $\mathcal{V}_G$ and $\mathcal{E}_G$ to denote its node set and edge set respectively. We call $\mathbf{v}=(v_1,\ldots, v_k)\in\mathcal{V}_G^k$ a \emph{$k$-tuple of nodes}, or simply a \emph{$k$-tuple} in $G$. If $C\subseteq \mathcal{V}_G$, we let $G[C]$ be the subgraph of $G$ induced by $C$.

Given $G\in\mathcal{G}$ and $u,v\in\mathcal{V}_G$, we use $d(u,v)$ to denote the shortest-path distance between $u$ and $v$. The set of \emph{$k$-th hop neighbors} of $u\in\mathcal{V}_G$ is defined as $\mathcal{N}_k(u)=\{v\in\mathcal{V}_G:d(u,v)=k\}$. We also denote $\mathcal{N}_{\leqslant k}(u)=\bigcup_{i=0}^k\mathcal{N}_i(u)$ and $\mathcal{N}(u)=\mathcal{N}_1(u)$.

\subsection{Isomorphisms and homomorphisms}

Let $G,H\in\mathcal{G}$, $\mathbf{u}=(u_1,\ldots, u_k)\in\mathcal{V}_G^k$ and $\mathbf{v}=(v_1,\ldots, v_\ell)\in\mathcal{V}_H^\ell$. The two tuples $\mathbf{u}$ and $\mathbf{v}$ are said to have \emph{the same isomorphism type} if (i) $k=\ell$, (ii) $\forall i,j\in[k], u_i=u_j\Leftrightarrow v_i=v_j$, and (iii) $\forall i,j\in[k], \{u_i,u_j\}\in\mathcal{E}_G\Leftrightarrow \{v_i,v_j\}\in\mathcal{E}_H$. The above definition gives rise to an equivalence relation between two arbitrary tuples (which can lie in two different graphs and have different lengths). Therefore, we may use the symbol $\mathrm{atp}(\mathbf{u})$ to refer to the equivalence class induced by the above relation, to which $\mathbf{u}$ belongs; namely, $\mathrm{atp}(\mathbf{u})=\mathrm{atp}(\mathbf{v})$ if and only if $\mathbf{u}$ and $\mathbf{v}$ have the same isomorphism type.

A \emph{homomorphism} from graph $G$ to graph $H$ is a mapping $h:\mathcal{V}_G\rightarrow \mathcal{V}_H$ such that for every $\{u,v\}\in\mathcal{E}_G$, we have $\{h(u),h(v)\}\in\mathcal{E}_H$. $H$ is called a \emph{homomorphism image} of $G$ if there exists a homomorphism $h$ from $G$ to $H$ satisfying (i) $h$ is surjective, and (ii) for every $u,v\in\mathcal{V}_G$ such that $\{u,v\}\notin\mathcal{E}_G$, we have $\{h(u),h(v)\}\notin\mathcal{E}_H$. We denote the set of all homomorphism images of $G$ (up to isomorphism) as $\mathrm{Spasm}(G)$.

Given two graphs $F,G\in\mathcal{G}$, we use $C(F,G)$ to denote the number of different subgraphs of $G$ that are isomorphic to $F$, and use $\mathrm{Hom}(F,G)$ to denote the number of different homomorphisms from $F$ to $G$. Further given $\mathbf{u}=(u_1,\ldots, u_k)\in\mathcal{V}_F^k$ and $\mathbf{v}=(v_1,\ldots, v_k)\in\mathcal{V}_G^k$, we use $\mathrm{Hom}(F^\mathbf{u},G^\mathbf{v})$ or $\mathrm{Hom}(F^{u_1,\ldots, u_k},G^{v_1,\ldots, v_k})$ to denote the number of different homomorphisms $h$ from $F$ to $G$ that satisfy $h(u_i)=v_i$, for each $i=1,\ldots, k$.

\subsection{Invariant and equivariant sets}

Let $G,H\in\mathcal{G}$ be two graphs and $f:\mathcal{V}_G\rightarrow \mathcal{V}_H$ be a mapping. If $\mathbf{v}=(v_1,\ldots, v_k)\in\mathcal{V}_G^k$, we adopt the shorthand $f(\mathbf{v})=(f(v_1),\ldots, f(v_k))\in\mathcal{V}_H^k$. Similarly, if $\mathcal{A}\subseteq\mathcal{V}_G^k$ is a set of $k$-tuples in $G$, we denote $f(\mathcal{A})=\{f(\mathbf{v}):\mathbf{v}\in\mathcal{A}\}\subseteq\mathcal{V}_H^k$. We now define \emph{invariant} and \emph{equivariant sets of $k$-tuples}.

\begin{definition}[Invariant set]
    Let $\mathcal{A}(\cdot)$ be a function that maps every $G\in\mathcal{G}$ to a set of $k$-tuples in $G$. Namely, $\forall G\in\mathcal{G}$, $\mathcal{A}(G)\subseteq\mathcal{V}_G^k$. The function $\mathcal{A}(\cdot)$ is said to be \textbf{$\bm{k}$-invariant} if the following holds: for any graph $H$ isomorphic to $G$, with $f:\mathcal{V}_G\rightarrow \mathcal{V}_H$ being an isomorphism, we have $f(\mathcal{A}(G))=\mathcal{A}(H)$. If $\mathcal{A}(\cdot)$ is $k$-invariant, $\mathcal{A}(G)\subseteq\mathcal{V}_G^k$ is said to be an \textbf{invariant set of $\bm{k}$-tuples}, or simply a \textbf{$\bm{k}$-invariant set} of $G$, given any $G\in\mathcal{G}$.
\end{definition}

For example, one can easily verify that $\{(u,v)\in\mathcal{V}_G^2:\{u,v\}\in\mathcal{E}_G\}$ and $\{(u,v)\in\mathcal{V}_G^2:d(u,v)\leqslant 2\}$ are both 2-invariant sets of $G$, given any $G\in\mathcal{G}$.

\begin{definition}[Equivariant set]
    Let $\mathcal{A}(\cdot, \cdot)$ be a function that maps every $G\in\mathcal{G}$ along with an $\ell$-tuple $\mathbf{u}\in\mathcal{V}_G^\ell$ to a set of $k$-tuples in $G$. Namely, $\forall G\in\mathcal{G}$ and $\mathbf{u}\in\mathcal{V}_G^\ell$, $\mathcal{A}(G,\mathbf{u})\subseteq\mathcal{V}_G^k$. The function $\mathcal{A}(\cdot,\cdot)$ is said to be \textbf{$\bm{(\ell,k)}$-equivariant} if the following holds: for any graph $H$ isomorphic to $G$, with $f:\mathcal{V}_G\rightarrow\mathcal{V}_H$ being an isomorphism, we have $f(\mathcal{A}(G,\mathbf{u}))=\mathcal{A}(H,f(\mathbf{u}))$. If $\mathcal{A}(\cdot,\cdot)$ is $(\ell,k)$-equivariant, $\mathcal{A}(G,\mathbf{u})\subseteq\mathcal{V}_G^k$ is said to be an \textbf{equivariant set of $\bm{k}$-tuples}, or simply an \textbf{$\bm{(\ell,k)}$-equivariant set} of $G$ with respect to $\mathbf{u}$, given any $G\in\mathcal{G}$ and $\mathbf{u}\in\mathcal{V}_G^\ell$.
\end{definition}

For example, given any graph $G$ and $(u,v)\in\mathcal{V}_G^2$, $\{w:d(u,w)\leqslant 2\text{ and }d(w,v)\leqslant 2\}$ is a $(2,1)$-equivariant set of $G$, with respect to $(u,v)$.

\section{Related works}\label{sect:rel_work}

\paragraph{Comparisons with \citep{dell2018lovasz} and \citep{zhang2024weisfeilerlehman}.} Prior to our work, \citet{dell2018lovasz} and \citet{zhang2024weisfeilerlehman} have studied the homomorphism counting power of some variants of Weisfeiler-Leman algorithms. \citet{dell2018lovasz} shows that indistinguishability of two graphs $G$ and $H$ by $k$-FWL is equivalent to $\mathrm{Hom}(F,G)=\mathrm{Hom}(F,H)$ for all graphs $F$ with tree-width at most $k$. \citet{zhang2024weisfeilerlehman} further extends the result of \citep{dell2018lovasz} to three other variants of WL algorithms, namely subgraph $k$-WL, local $k$-WL and local 2-FWL, using the language of nested ear decompositions. However, even \citep{zhang2024weisfeilerlehman} has \textbf{not completely determined the homomorphism counting power of local $\bm{k}$-FWL, for the case of $\bm{k>2}$}; \citep{zhang2024weisfeilerlehman} also leaves the question open whether the so-called \emph{homomorphism expressivity} exists for any algorithms that adopt a color-refinement paradigm. Our work fills the gap in \citep{zhang2024weisfeilerlehman} regarding the homomorphism counting power of local $k$-FWL with $k>2$, and provides an affirmative answer to the latter question for the case of GFWL algorithms. 

Our work is also methodologically very different from \citep{dell2018lovasz} and \citep{zhang2024weisfeilerlehman}. The proof of our result no longer depends on concepts such as tree decompositions or nested ear decompositions, and is fully stated in the language of pebble games. Despite its relative conceptual simplicity, our result is much more general than those in \citep{dell2018lovasz} and \citep{zhang2024weisfeilerlehman}, indicating that our approach (using pebble games) represents a better perspective to tackle the homomorphism counting problem.

\paragraph{Pebble games on graphs.} The Ehrenfeucht-Fra\"iss\'e games, named after \citet{ehrenfeucht1961application} and \citet{fraisse1954quelques}, have been used to analyze the $k$-FWL algorithms in \citep{cai1992optimal, grohe2015pebble}. More recent works \citep{zhang2023complete, zhao2022practical} generalize the Ehrenfeucht-Fra\"iss\'e games to provide game-theoretic characterizations for other variants of WL algorithms.

The Cops-Robber games, initially appearing as an instance of \emph{graph searching games}, are intensively studied by \citet{seymour1993graph}. The original version of the game includes $k$ Cops trying to trap a Robber on a graph $G$, where Robber is only capable of moving along the edges to a Cop-free node. The central result of \citep{seymour1993graph} is a vital property of the game: if Cops can guarantee to win the game, they can win \emph{monotonically}, meaning that they can ensure Robber never return to a node they have previously expelled Robber from. One corollary of the property is that Cops can win the Cops-Robber game on $G$ if and only if $k\geqslant \text{tree-width}(G)+1$. The standard textbook \citep{citeulike:395714} on graph theory provides a simpler proof for the latter result. Several recent works \citep{neuen2023homomorphism, zhang2023complete, zhang2024weisfeilerlehman} have revived the Cops-Robber games by relating them to F\"urer graphs \citep{cai1992optimal}, which are standard testers for expressive power. However, none of the existing works have directly examined \textbf{the relationship between Cops-Robber games and the homomorphism counting power of WL algorithms}. As is presented in our main theorem (Theorem \ref{main_theorem}), such a relationship is surprisingly simple and beautiful.

We also remark that Appendix A.1 of \citep{zhang2024weisfeilerlehman} provides a detailed review of existing approaches to the design of expressive GNNs, which is closely related to our work.

\section{Results}\label{section_prelim}

\subsection{Generalized folklore Weisfeiler-Leman algorithms}

In this subsection, we present the definition of \emph{generalized folklore Weisfeiler-Leman algorithms}, which extend $(k,t)$-FWL+ defined in~\citep{feng2023towards}. Before embarking on the detailed discussion, we introduce a few notations.

\begin{definition}[Concatenation]
    Let $\mathbf{v}=(v_1,\ldots, v_k)\in\mathcal{V}_G^k$ and $\mathbf{u}=(u_1,\ldots, u_\ell)\in\mathcal{V}_G^\ell$. The $(k+\ell)$-tuple $(v_1,\ldots, v_k,u_1,\ldots, u_\ell)\in\mathcal{V}_G^{k+\ell}$ is called the \textbf{concatenation} of $\mathbf{v}$ and $\mathbf{u}$, denoted as $(\mathbf{v},\mathbf{u})$.
\end{definition}

The following definition follows that of \emph{neighborhood tuples} introduced in~\citep{feng2023towards}, but is slightly different.

\begin{definition}[Replacement]
    Let $\mathbf{v}=(v_1,\ldots, v_k)\in\mathcal{V}_G^k$ and $\mathbf{u}=(u_1,\ldots, u_\ell)\in\mathcal{V}_G^\ell$. Denote $v_{k+1}=u_1,\ldots, v_{k+\ell}=u_\ell$. For any $k$ indices $r_1,\ldots, r_k\in[k+\ell]$ with $r_1<r_2<\cdots<r_k$, the $k$-tuple
    \begin{gather}
        \mathrm{REPLACE}(\mathbf{v},\mathbf{u};r_1,\ldots, r_k)=(v_{r_1},\ldots, v_{r_k})\in\mathcal{V}_G^k
    \end{gather}
    is called a \textbf{replacement of $\mathbf{v}$ by $\mathbf{u}$}.
\end{definition}

The most basic operation of generalized folklore Weisfeiler-Leman algorithms is \emph{aggregation}, which we describe below.

\begin{definition}
    Let $\mathcal{A}\subseteq\mathcal{V}_G^k$ and $\ell<k$. For any $\mathbf{u}=(u_1,\ldots, u_\ell)\in\mathcal{V}_G^\ell$, denote
    \begin{gather}
        \mathcal{A}\backslash\mathbf{u} = \{(w_1,\ldots, w_{k-\ell})\in\mathcal{V}_G^{k-\ell}:(u_1,\ldots, u_\ell, w_1,\ldots, w_{k-\ell})\in\mathcal{A}\}.
    \end{gather}
    Let $W(\cdot)$ be a function defined on $\mathcal{A}$. Define $\mathcal{A}'\subseteq\mathcal{V}_G^\ell$ and a new function $W'(\cdot)$ on $\mathcal{A}'$ as following:
    \begin{align}
        \mathcal{A}'&=\{\mathbf{u}\in\mathcal{V}_G^\ell:\mathcal{A}\backslash\mathbf{u}\ne\varnothing\},\label{set-aggr-reduce}\\
        W'(\mathbf{u})&=\mathcal{H}\left(\lbbr W(\mathbf{v}): \mathbf{v}=(\mathbf{u},\mathbf{w}), \text{ for some }\mathbf{w}\in\mathcal{A}\backslash\mathbf{u}\rbbr\right),\quad \forall \mathbf{u}\in\mathcal{A}',\label{aggregation-reduce}
    \end{align}
    where $\mathcal{H}(\cdot)$ is an injective hashing function operating on multisets. The above procedure producing $\mathcal{A}'$ and $W'(\cdot)$ is called an \textbf{aggregation operation on $\bm{\mathcal{A}}$ and $\bm{W(\cdot)}$}. We adopt the notations
    \begin{align}
        \mathcal{A'}&=\mathrm{AGGR}_\ell^\mathcal{H}(\mathcal{A}),\\
        W'(\mathbf{u})&=\mathrm{AGGR}_\ell^\mathcal{H}(\mathbf{u};W(\cdot),\mathcal{A}),\quad \forall \mathbf{u}\in\mathcal{A}'
    \end{align}
    as mere shorthands for equations \eqref{set-aggr-reduce} and \eqref{aggregation-reduce}. We often omit the superscript $\mathcal{H}$ in $\mathrm{AGGR}_\ell^\mathcal{H}$ if we are not interested in the specific choice of $\mathcal{H}(\cdot)$. We also note that when $\ell=0$, equation \eqref{aggregation-reduce} yields a single value $W'=\mathcal{H}\left(\lbbr W(\mathbf{v}): \mathbf{v}\in\mathcal{A}\rbbr\right)$ with no $\mathbf{u}$ dependence. 
\end{definition}

We are now ready to give a formal description of generalized folklore Weisfeiler-Leman (GFWL) algorithms. Any specific instance of the algorithm depends on several ``hyperparameters'', which we list below.

\begin{assumption}\label{assump}
    Let $k,t\in\mathbb{N}^*$, $0=i_0<i_1<i_2<\cdots<i_N=k$ and $0=j_0<j_1<j_2<\cdots<j_M=t$. Further let $\mathcal{R}(\cdot)$ be a $k$-invariant function, and let $\mathcal{F}(\cdot, \cdot)$ be a $(k,t)$-equivariant function.
\end{assumption}

\begin{definition}[Generalized folklore Weisfeiler-Leman algorithms]\label{gfwl_def}
    The algorithm computes $W(G)$ for any graph $G\in\mathcal{G}$ by executing the following steps:
    \begin{itemize}
        \item \textbf{Initialization.} Assign an initial color $W^{(0)}(\mathbf{v})$ to every $\mathbf{v}\in\mathcal{R}(G)$. We require that
        \begin{gather}
            W^{(0)}(\mathbf{v})=\Phi(\mathrm{atp}(\mathbf{v})),
        \end{gather}
        where $\Phi(\cdot)$ is an injective function.
        \item \textbf{Update.} An update step includes \emph{message computation} and \emph{aggregation}. In the following, we let $W^{(T)}(\mathbf{v})$ be the color of $\mathbf{v}\in\mathcal{R}(G)$ at the $T$-th iteration, for $T=0,1,\ldots$.
        \begin{itemize}
            \item[$\circ$] \textbf{Message computation.} For each $\mathbf{v}\in\mathcal{R}(G)$ and $\mathbf{u}\in\mathcal{F}(G,\mathbf{v})$, there are $C=\binom{k+t}{k}$ different replacements of $\mathbf{v}$ by $\mathbf{u}$, denoted as
            \begin{gather}
                \mathbf{v}_c(\mathbf{u})=\mathrm{REPLACE}(\mathbf{v},\mathbf{u};r_1^{(c)},\ldots, r_k^{(c)}),\quad c=1,\ldots, C.
            \end{gather}
            We sort $\{\mathbf{v}_c(\mathbf{u}):c\in[C]\}$ according to the lexicographic order of $(r_1^{(c)},\ldots, r_k^{(c)})$. Compute the message from $\mathbf{u}$ to $\mathbf{v}$ as following:\footnote{We implicitly assume that each $\mathbf{v}_c(\mathbf{u})\in\mathcal{R}(G)$, for any $\mathbf{v}\in\mathcal{R}(G)$, $\mathbf{u}\in\mathcal{F}(G,\mathbf{v})$ and $c\in\left[C\right]$. }
            \begin{gather}
                \mathrm{MSG}^{(T)}(\mathbf{u};\mathbf{v}) = \left(W^{(T)}(\mathbf{v}_1(\mathbf{u})),\ldots, W^{(T)}(\mathbf{v}_C(\mathbf{u}))\right).\label{message_definition}
            \end{gather}
            Thus, $\mathrm{MSG}^{(T)}(\cdot;\mathbf{v})$ is a color-vector-valued function on $\mathcal{F}(G,\mathbf{v})$.
            \item[$\circ$] \textbf{Aggregation.} For every $\mathbf{v}\in\mathcal{R}(G)$, perform $M$ consecutive aggregation operations on $\mathcal{F}(G,\mathbf{v})$ and $\mathrm{MSG}^{(T)}(\cdot;\mathbf{v})$: setting $\mathcal{F}_M(G,\mathbf{v})=\mathcal{F}(G,\mathbf{v})$ and $\mathrm{MSG}^{(T)}_M(\cdot;\mathbf{v}) = \mathrm{MSG}^{(T)}(\cdot;\mathbf{v})$, we let
            \begin{align}
                \mathcal{F}_{m-1}(G,\mathbf{v}) &= \mathrm{AGGR}_{j_{m-1}}(\mathcal{F}_m(G,\mathbf{v}))\subseteq\mathcal{V}_G^{j_{m-1}},\label{equivariant_set}\\
                \mathrm{MSG}^{(T)}_{m-1}(\mathbf{u};\mathbf{v}) &= \mathrm{AGGR}_{j_{m-1}}(\mathbf{u};\mathrm{MSG}^{(T)}_m(\cdot;\mathbf{v}),\mathcal{F}_m(G,\mathbf{v})),\quad \mathbf{u}\in\mathcal{F}_{m-1}(G,\mathbf{v}),
            \end{align}
            for $m=M,M-1,\ldots, 1$.\footnote{For aggregation operations at different steps (i.e., $\mathrm{AGGR}_{j_{m-1}}$ with different $m$ values), different $\mathcal{H}(\cdot)$ functions can be used. A similar remark applies to the pooling step.} Since $j_0=0$, the last aggregation operation yields $\mathrm{MSG}^{(T)}_0(\mathbf{v})$ which is independent of $\mathbf{u}$. We let
            \begin{gather}
                W^{(T+1)}(\mathbf{v}) = \mathrm{MSG}^{(T)}_0(\mathbf{v}).
            \end{gather}
        \end{itemize}
        \item \textbf{Pooling.} Repeat the update step until the colors $W^{(T)}(\mathbf{v}),\mathbf{v}\in\mathcal{R}(G)$ go stable. Denote the stable coloring of $\mathbf{v}$ as $W^{(\infty)}(\mathbf{v})$. Perform $N$ consecutive aggregation operations on $\mathcal{R}(G)$ and $W^{(\infty)}(\cdot)$: setting $\mathcal{R}_N(G)=\mathcal{R}(G)$ and $W_N^{(\infty)}(\cdot)=W^{(\infty)}(\cdot)$, we let
        \begin{align}
            \mathcal{R}_{n-1}(G) &= \mathrm{AGGR}_{i_{n-1}}(\mathcal{R}_n(G))\subseteq\mathcal{V}_G^{i_{n-1}},\label{invariant_set}\\
            W_{n-1}^{(\infty)}(\mathbf{v}) &= \mathrm{AGGR}_{i_{n-1}}(\mathbf{v}; W_n^{(\infty)}(\cdot),\mathcal{R}_n(G)),\quad \mathbf{v}\in\mathcal{R}_{n-1}(G),\label{invariant_aggregation}
        \end{align}
        for $n=N,N-1,\ldots, 1$. Since $i_0=0$, we eventually get a single color $W^{(\infty)}_0$ independent of $\mathbf{v}$. We let $W(G) = W^{(\infty)}_0$.
    \end{itemize}
\end{definition}

One can see that if we set $N=k,M=t$ and $(i_0,i_1,\ldots, i_N)=(0,1,\ldots, k),(j_0,j_1,\ldots, j_M)=(0,1,\ldots, t)$ in Assumption \ref{assump}, our Definition \ref{gfwl_def} gives a $(k,t)$-FWL+ instance as defined in~\citep{feng2023towards}. We point out that many other variants of WL algorithms~\citep{zhang2023complete, zhang2024weisfeilerlehman, zhou2024distance} can be unified into the GFWL framework, as is shown in Appendix \ref{example_wl}.

\subsection{Pebble games}

In this paper, we consider two types of pebble games: the \emph{Ehrenfeucht-Fra\"iss\'e game}~\citep{cai1992optimal, dell2018lovasz, ehrenfeucht1961application, fraisse1954quelques, grohe2015pebble} and the \emph{Cops-Robber game}~\citep{neuen2023homomorphism, seymour1993graph, zhang2023complete, zhang2024weisfeilerlehman}. We assume that Assumption \ref{assump} holds throughout this subsection.

\begin{definition}[Ehrenfeucht-Fra\"iss\'e game] 
    Assume there are a total of $(k+t)$ pairs of pebbles $(p_1,q_1),\ldots, (p_{k+t},$ $q_{k+t})$, all initially off the table. We call the pebbles $(p_1, q_1),\ldots, (p_k,q_k)$ \textbf{main pebbles}, and the pebbles $(p_{k+1}, q_{k+1}),$ $\ldots, (p_{k+t}, q_{k+t})$ \textbf{auxiliary pebbles}. 
    
    Let $G,H\in\mathcal{G}$ be two graphs placed on the table. Two players (named \textbf{Spoiler} and \textbf{Duplicator} respectively)\footnote{We will use masculine pronouns for Spoiler, and feminine pronouns for Duplicator.} play the \textbf{Ehrenfeucht-Fra\"iss\'e game} on $G$ and $H$ as following:
    
    \begin{itemize}
        \item The two players first play an \textbf{initialization round} with $N$ \textbf{putting moves}, numbered by $n=1,\ldots, N$. The game rule guarantees that before the $n$-th putting move begins, the pebbles $p_1,\ldots, p_{i_{n-1}}$ occupy an $i_{n-1}$-tuple $\mathbf{u}_{n-1}^{(0)}$ of $G$, while the pebbles $q_1,\ldots, q_{i_{n-1}}$ occupy an $i_{n-1}$-tuple $\mathbf{v}_{n-1}^{(0)}$ of $H$.\footnote{Before the first ($n=1$) putting move, there should be no pebbles on either graph, since $i_0=0$.} The $n$-th putting move includes the following steps:
        \begin{enumerate}
            \item Duplicator chooses a bijection $f:\mathcal{R}_n(G)\backslash \mathbf{u}^{(0)}_{n-1}\rightarrow \mathcal{R}_n(H)\backslash\mathbf{v}^{(0)}_{n-1}$, where $\mathcal{R}_n(\cdot)$ is defined in equation \eqref{invariant_set}.
            \item Spoiler chooses $\Delta\mathbf{u}^{(0)}_n\in \mathcal{R}_n(G)\backslash \mathbf{u}^{(0)}_{n-1}$. Let $\Delta\mathbf{v}^{(0)}_n=f(\Delta\mathbf{u}^{(0)}_n)$. He then puts pebbles $(p_{i_{n-1}+1},p_{i_{n-1}+2},\ldots, p_{i_n})$ at $\Delta\mathbf{u}^{(0)}_n$, and also puts $(q_{i_{n-1}+1},q_{i_{n-1}+2},\ldots, q_{i_n})$ at $\Delta\mathbf{v}^{(0)}_n$.
        \end{enumerate}
        After the $n$-th putting move, $p_1,\ldots, p_{i_n}$ occupy $\mathbf{u}^{(0)}_n=(\mathbf{u}^{(0)}_{n-1},\Delta \mathbf{u}^{(0)}_n)\in\mathcal{R}_n(G)$, and $q_1,\ldots, q_{i_n}$ occupy $\mathbf{v}^{(0)}_n=(\mathbf{v}^{(0)}_{n-1},\Delta\mathbf{v}^{(0)}_n)\in\mathcal{R}_n(H)$. Especially, after all $N$ putting moves, $p_1,\ldots, p_k$ should occupy some $\mathbf{u}^{(0)}_N\in\mathcal{R}(G)$, while $q_1,\ldots, q_k$ occupy some $\mathbf{v}^{(0)}_N\in\mathcal{R}(H)$.
        \item After the initialization round, the two players repeatedly play the \textbf{update round}, each with $M$ \textbf{putting moves} (numbered by $m=1,\ldots, M$) and one \textbf{removing move}. 
        
        Throughout the $M$ putting moves of the $T$-th update round, the main pebbles $p_1,\ldots, p_k$ and $q_1,\ldots, q_k$ stay unmoved at $\mathbf{u}^{(T-1)}_N$ and $\mathbf{v}^{(T-1)}_N$, respectively. The game rule guarantees that before the $m$-th putting move begins, the auxiliary pebbles $p_{k+1},\ldots, p_{k+j_{m-1}}$ occupy a $j_{m-1}$-tuple $\mathbf{u}'^{(T)}_{m-1}$ of $G$, while $q_{k+1},\ldots, q_{k+j_{m-1}}$ occupy a $j_{m-1}$-tuple $\mathbf{v}'^{(T)}_{m-1}$ of $H$. The $m$-th putting move includes the following steps:
        \begin{enumerate}
            \item Duplicator chooses a bijection $f:\mathcal{F}_m(G,\mathbf{u}^{(T-1)}_N)\backslash \mathbf{u}'^{(T)}_{m-1}\rightarrow \mathcal{F}_m(H,\mathbf{v}^{(T-1)}_N)\backslash \mathbf{v}'^{(T)}_{m-1}$, where $\mathcal{F}_m(\cdot,\cdot)$ is defined in equation \eqref{equivariant_set}.
            \item Spoiler chooses $\Delta\mathbf{u}'^{(T)}_m\in \mathcal{F}_m(G,\mathbf{u}^{(T-1)}_N)\backslash \mathbf{u}'^{(T)}_{m-1}$. Let $\Delta\mathbf{v}'^{(T)}_m=f(\Delta\mathbf{u}'^{(T)}_m)$. He then puts pebbles $(p_{k+j_{m-1}+1},p_{k+j_{m-1}+2},\ldots, p_{k+j_m})$ at $\Delta\mathbf{u}'^{(T)}_m$, and also puts $(q_{k+j_{m-1}+1},q_{k+j_{m-1}+2},\ldots, q_{k+j_m})$ at $\Delta\mathbf{v}'^{(T)}_m$.
        \end{enumerate}
        After the $m$-th putting move, the pebbles $p_{k+1},\ldots, p_{k+j_m}$ occupy $\mathbf{u}'^{(T)}_m=(\mathbf{u}'^{(T)}_{m-1},\Delta \mathbf{u}'^{(T)}_m)\in\mathcal{F}_m(G,\mathbf{u}^{(T-1)}_N)$, and pebbles $q_{k+1},\ldots, q_{k+j_m}$ occupy $\mathbf{v}'^{(T)}_m=(\mathbf{v}'^{(T)}_{m-1},\Delta \mathbf{v}'^{(T)}_m)\in\mathcal{F}_m(H,\mathbf{v}^{(T-1)}_N)$.
        
        After all $M$ putting moves (now we have $p_1,\ldots, p_{k+t}$ and $q_1,\ldots, q_{k+t}$ located on $G$ and $H$ perspectively), the removing move goes as follows. Spoiler arbitrarily chooses $k$ indices $r_1,\ldots, r_k\in[k+t]$ with $r_1<\cdots<r_k$. He then removes all pebbles on $G$ other than $p_{r_1},\ldots, p_{r_k}$, and relabels the pebbles such that $p_{r_s}$ becomes $p_s$, for $s=1,\ldots, k$. Accordingly, Spoiler also removes all pebbles on $H$ other than $q_{r_1},\ldots, q_{r_k}$, and relabels $q_{r_s}$ to $q_s$, for $s=1,\ldots, k$. Finally, he lets $\mathbf{u}^{(T)}_N$ and $\mathbf{v}^{(T)}_N$ be the $k$-tuples occupied by (pebbles now labeled as) $p_1,\ldots, p_k$ and $q_1,\ldots, q_k$, respectively.
    \end{itemize}
    
    Spoiler \textbf{wins the game} if either of the following happens: (i) Duplicator fails to find a bijection $f$ in a putting move; (ii) when a (putting or removing) move ends, $\mathrm{atp}(\mathbf{u})\ne\mathrm{atp}(\mathbf{v})$, where $\mathbf{u}$ and $\mathbf{v}$ are the tuples of nodes occupied by pebbles on $G$ and $H$ (ordered by the indices of pebbles) at the end of the move. Duplicator \textbf{wins the game} if Spoiler cannot win the game after any number of moves.
\end{definition}

\begin{definition}[Cops-Robber game]\label{cr-def}
    Assume there are a total of $(k+t)$~pebbles $p_1,\ldots, p_{k+t}$, all initially off the table. We call the pebbles $p_1,\ldots, p_k$ \textbf{main pebbles}, and the pebbles $p_{k+1},\ldots, p_{k+t}$ \textbf{auxiliary pebbles}.

    Let $G\in\mathcal{G}$ be a graph placed on the table. Two players (named \textbf{Cops} and \textbf{Robber} respectively)\footnote{We will use plural verbs and pronouns for Cops, although ``Cops'' is indeed an individual player. For Robber, we will use masculine pronouns.} play the \textbf{Cops-Robber game} on $G$ as following:

    \begin{itemize}
        \item Throughout the game, Robber maintains a subset $C$ of $\mathcal{V}_G$ such that $G[C]$ is connected. Before the game starts, Robber chooses $C$ as a connected component\footnote{A \emph{connected component} of $G$ is an equivalence class on $\mathcal{V}_G$ under the reachability relation.} of $G$.
        \item The two players first play an \textbf{initialization round} with $N$ \textbf{guarding moves}, numbered by $n=1,\ldots, N$. The game rule guarantees that before the $n$-th guarding move begins, the pebbles $p_1,\ldots, p_{i_{n-1}}$ occupy an $i_{n-1}$-tuple $\mathbf{u}^{(0)}_{n-1}$ of $G$. The $n$-th guarding move goes as follows:
        \begin{enumerate}
            \item Cops choose $\Delta\mathbf{u}^{(0)}_n\in \mathcal{R}_n(G)\backslash \mathbf{u}^{(0)}_{n-1}$, and put pebbles $(p_{i_{n-1}+1},p_{i_{n-1}+2},\ldots, p_{i_n})$ at $\Delta\mathbf{u}^{(0)}_n$. After Cops take the action, $p_1,\ldots, p_{i_n}$ should occupy $\mathbf{u}^{(0)}_n=(\mathbf{u}^{(0)}_{n-1},\Delta\mathbf{u}^{(0)}_n)\in\mathcal{R}_n(G)$.
            \item Let $X=\{u\in\mathcal{V}_G:u \text{ is occupied by a pebble}\}$. Robber chooses a connected component $C'$ of $G-X$ that satisfies $C'\subseteq C$, and updates his subset $C$ to $C'$.
        \end{enumerate}
        After all $N$ guarding moves, the main pebbles $p_1,\ldots, p_k$ should occupy some $\mathbf{u}^{(0)}_N\in\mathcal{R}(G)$.
        \item After the initialization round, the two players repeatedly play the \textbf{update round}, each with $M$ \textbf{guarding moves} (numbered by $m=1,\ldots, M$) and one \textbf{unguarding move}. 

        Throughout the $M$ guarding moves of the $T$-th update round, $p_1,\ldots, p_k$ stay unmoved at $\mathbf{u}^{(T-1)}_N$. The game rule guarantees that before the $m$-th guarding move begins, the auxiliary pebbles $p_{k+1},\ldots, p_{k+j_{m-1}}$ occupy a $j_{m-1}$-tuple $\mathbf{u}'^{(T)}_{m-1}$ of $G$. The $m$-th guarding move goes as follows:
        \begin{enumerate}
            \item Cops choose $\Delta\mathbf{u}'^{(T)}_m\in \mathcal{F}_m(G,\mathbf{u}^{(T-1)}_N)\backslash \mathbf{u}'^{(T)}_{m-1}$, and put pebbles $(p_{k+j_{m-1}+1},$ $p_{k+j_{m-1}+2},\ldots, p_{k+j_m})$ at $\Delta\mathbf{u}'^{(T)}_m$. After Cops take the action, $p_{k+1},\ldots, p_{k+j_m}$ occupy $\mathbf{u}'^{(T)}_m=(\mathbf{u}'^{(T)}_{m-1},\Delta \mathbf{u}'^{(T)}_m)$ $\in\mathcal{F}_m(G,\mathbf{u}^{(T-1)}_N)$.
            \item Let $X=\{u\in\mathcal{V}_G:u \text{ is occupied by a pebble}\}$. Robber chooses a connected component $C'$ of $G-X$ that satisfies $C'\subseteq C$, and updates his subset $C$ to $C'$.
        \end{enumerate}

        After the $M$ guarding moves (now all $(k+t)$ pebbles should be on $G$), the unguarding move goes as follows. Cops arbitrarily choose $k$ indices $r_1,\ldots, r_k\in[k+t]$ with $r_1<\cdots<r_k$. They then remove all pebbles on $G$ other than $p_{r_1},\ldots, p_{r_k}$, and relabel the pebbles such that $p_{r_s}$ becomes $p_s$, for $s=1,\ldots, k$. After Cops are done, let $X=\{u\in\mathcal{V}_G:u \text{ is occupied by a pebble}\}$. Robber chooses a connected component $C'$ of $G-X$ that satisfies $C'\supseteq C$, and updates his subset $C$ to $C'$. Finally, $\mathbf{u}^{(T)}_N$ is set to the new tuple occupied by (pebbles now labeled as) $p_1,\ldots, p_k$.
    \end{itemize}

    If Robber fails to find a connected component $C'$ according to the rule in some move, he is said to lose that move. Cops \textbf{win the game} if they can make Robber lose in some move. Robber \textbf{wins the game} if Cops can never win the game.
\end{definition}

\subsection{Homomorphism counting power of GFWL algorithms}

In this subsection, we present our main result regarding the homomorphism counting power of generalized folklore Weisfeiler-Leman algorithms.

\begin{definition}
    Let $\mathcal{A}(\cdot)$ be a $k$-invariant function. $\mathcal{A}(\cdot)$ is said to be \textbf{closed under homomorphisms} if for any $G,H\in\mathcal{G}$ with a homomorphism $h$ from $G$ to $H$, holds $h(\mathcal{A}(G))\subseteq\mathcal{A}(H)$.
\end{definition}

\begin{definition}
    Let $\mathcal{A}(\cdot, \cdot)$ be an $(\ell,k)$-equivariant function. $\mathcal{A}(\cdot,\cdot)$ is said to be \textbf{closed under homomorphisms} if for any $G,H\in\mathcal{G}$ with a homomorphism $h$ from $G$ to $H$, and for any $\mathbf{u}\in\mathcal{V}_G^\ell$, holds $h(\mathcal{A}(G,\mathbf{u}))\subseteq\mathcal{A}(H,h(\mathbf{u}))$.
\end{definition}

Our main theoretical result is the following theorem. It is assumed that Assumption \ref{assump} holds for the theorem.

\begin{theorem}\label{main_theorem}
    Let $F\in\mathcal{G}$. Assume that $\mathcal{R}(\cdot)$ and $\mathcal{F}(\cdot,\cdot)$ in Assumption \ref{assump} are both closed under homomorphisms. Then the following two statements are equivalent:\\
    \textbf{(a).} $G$ and $H$ get different colors $W(G)$ and $W(H)$ from the Weisfeiler-Leman algorithm;\\
    \textbf{(b).} $\exists F\in\mathcal{G}$ such that Cops win the Cops-Robber game on $F$ and that $\mathrm{Hom}(F,G)\ne\mathrm{Hom}(F,H)$.
\end{theorem}

Informally speaking, Theorem~\ref{main_theorem} determines that the set
\begin{gather}
\{F\in\mathcal{G}:\text{Cops win the Cops-Robber game on }F\}    
\end{gather}
is exactly the set of query graphs whose homomorphisms a certain GFWL algorithm (defined by hyperparameters in Assumption \ref{assump}) can count. The proof of Theorem~\ref{main_theorem} is split into two parts, each of which we summarize as a theorem below.

\begin{theorem}\label{theorem:weis-leman-ehren-frai}
    Let $G,H\in\mathcal{G}$. The following two statements are equivalent:\\
    \textbf{(a).} $G$ and $H$ get different colors $W(G)$ and $W(H)$ from the Weisfeiler-Leman algorithm;\\
    \textbf{(b).} Spoiler wins the Ehrenfeucht-Fra\"iss\'e game on $G$ and $H$.
\end{theorem}

\begin{theorem}\label{theorem:cops-robber-ehren-frai}
    Let $G,H\in\mathcal{G}$. Assume that $\mathcal{R}(\cdot)$ and $\mathcal{F}(\cdot,\cdot)$ are both closed under homomorphisms. Then the following two statements are equivalent:\\
    \textbf{(a).} Spoiler wins the Ehrenfeucht-Fra\"iss\'e game on $G$ and $H$;\\
    \textbf{(b).} $\exists F\in\mathcal{G}$ such that Cops win the Cops-Robber game on $F$ and that $\mathrm{Hom}(F,G)\ne\mathrm{Hom}(F,H)$.
\end{theorem}

The proofs of Theorems~\ref{theorem:weis-leman-ehren-frai} and~\ref{theorem:cops-robber-ehren-frai} are given in Appendices~\ref{app:proof1} and~\ref{app:proof2} respectively.

\section{Future directions}\label{sec:future}

In this paper, we propose generalized folklore Weisfeiler-Leman algorithms to unify a number of existing designs of powerful GNNs. Currently, our framework is only suitable to describe GNNs (or WL variants) of folklore Weisfeiler-Leman (FWL) type. Since FWL-type variants usually achieves higher expressive power at the same computational complexities, it is practically acceptable to consider their homomorphism counting power only. Nevertheless, the homomorphism counting power of other variants of WL algorithms, such as $k$-WL~\citep{morris2019weisfeiler}, sparse $k$-WL~\citep{morris2020weisfeiler} or subgraph WL~\citep{zhang2023complete}, remains an interesting theoretic question. A future direction is to generalize our discussion on the homomorphism counting power of GFWL algorithms to those other variants. This would entail some modifications in the definitions of GFWL algorithms, as well as the two types of pebble games. In principle, the generalization would be straightforward. 

Another more important question is the connection between homomorphism counting power and substructure counting power of WL variants. So far, only two works~\citep{neuen2023homomorphism, zhang2024weisfeilerlehman} have yielded satisfactory results for some specific WL variants, namely $k$-FWL, subgraph $k$-WL, local $k$-WL and local 2-FWL. For all those mentioned WL variants, their ability to count subgraph isomorphisms from a query graph $H$ is shown to be equivalent to the ability to count subgraph homomorphisms from \textbf{all graphs within $\bm{\mathrm{Spasm}(H)}$}. However, it still remains unknown whether such result can be generalized to the case of GFWL. The hard point to establish such equivalence lies in the necessity to prove the monotonicity of Cops-Robber game (Definition~\ref{cr-def}) that corresponds to an arbitrary GFWL instance. As we have reviewed in Section~\ref{sect:rel_work}, the monotonicity proof has been made for $k$-FWL~\citep{citeulike:395714, seymour1993graph}, and the other three variants listed above~\citep{zhang2024weisfeilerlehman}. The proof methodologies in those works cannot be easily generalized to the case of GFWL. Therefore, we leave the general monotonicity proof for GFWL, as well as the problem of determining the exact substructure counting power of GFWL, to future works.

\bibliography{references}
\bibliographystyle{plainnat}
\newpage
\appendix
\section{Examples of generalized Weisfeiler-Leman algorithms}\label{example_wl}

\begin{proposition}
    A GFWL instance with $t=1$, $N=1$, $M=1$, $(i_0,i_1)=(0,k)$, $(j_0,j_1)=(0,1)$, $\mathcal{R}(G)=\mathcal{V}_G^k$ and $\mathcal{F}(G,\mathbf{u})=\mathcal{V}_G$ for any $\mathbf{u}\in\mathcal{V}_G^k$ in Assumption~\ref{assump} gives an instance of $k$-FWL~\citep{cai1992optimal, maron2019provably}.
\end{proposition}

\begin{proposition}
    A GFWL instance with $t=1$, $N=1$, $M=1$, $(i_0,i_1)=(0,k)$, $(j_0,j_1)=(0,1)$, $\mathcal{R}(G)=\mathcal{V}_G^k$ and $\mathcal{F}(G,\mathbf{u})=\bigcup_{\ell=1}^k \mathcal{N}(u_\ell)$ for $\mathbf{u}=(u_1,\ldots, u_k)\in\mathcal{V}_G^k$ in Assumption~\ref{assump} gives an instance of local $k$-FWL~\citep{zhang2024weisfeilerlehman}. When $k=2$, this is exactly the SLFWL(2) proposed in~\citet{zhang2023complete}, which is strictly more powerful than all node-based subgraph GNNs (the latter defined by~\citet{frascaunderstanding2022}).
\end{proposition}

\begin{proposition}
    A GFWL instance with $k=2$, $t=1$, $N=1$, $M=1$, $(i_0,i_1)=(0,2)$, $(j_0,j_1)=(0,1)$, $\mathcal{R}(G)=\{(u,v)\in\mathcal{V}_G^2:d(u,v)\leqslant \delta\}$ and $\mathcal{F}(G,(u,v))=\{w\in\mathcal{V}_G:d(u,w)\leqslant \delta\text{ and } d(w,v)\leqslant \delta\}$ in Assumption~\ref{assump} gives an instance of $\delta$-DRFWL(2)~\citep{zhou2023distance}.
\end{proposition}

\begin{proposition}
    A GFWL instance with $N=k$, $M=t$, $(i_0,i_1,\ldots, i_N)=(0,1,\ldots, k)$ and $(j_0,j_1,\ldots, j_M)=(0,1,\ldots, t)$ in Assumption~\ref{assump} gives an instance of $(k,t)$-FWL+~\citep{feng2023towards}. Further with $\mathcal{R}(G)=\mathcal{V}_G^k$ and $\mathcal{F}(G,\mathbf{u})=\mathcal{V}_G^t$ for any $\mathbf{u}\in\mathcal{V}_G^k$, the resulting GFWL instance is an instance of $(k,t)$-FWL~\citep{feng2023towards}.
\end{proposition}

\section{Proof of Theorem~\ref{theorem:weis-leman-ehren-frai}}\label{app:proof1}

\begin{proof}
    \textbf{(a) $\bm{\Rightarrow}$ (b).} Since $W(G)\ne W(H)$, we have
    \begin{gather}
        W_0^{(\infty)}|_G \ne W_0^{(\infty)}|_H, \label{different_colors_init}
    \end{gather}
    where we have used the subscripts to disambiguate between the colors computed for graphs $G$ and $H$. Our proof relies on the following lemma.
    \begin{lemma}\label{theorem1_lemma1}
        Given equation \eqref{different_colors_init}, the following statement holds for each $n=1,2,\ldots, N$: if Duplicator has not lost after the $(n-1)$-th putting move of the initialization round, then either (i) Duplicator loses after the $n$-th putting move of the initialization round, or (ii) Spoiler can maintain that 
        \begin{gather}
            W_n^{(\infty)}(\mathbf{u}_n)|_G\ne W_n^{(\infty)}(\mathbf{v}_n)|_H
        \end{gather}
        after the $n$-th putting move of the initialization round.
    \end{lemma}
    \emph{Proof of Lemma \ref{theorem1_lemma1}.} We prove by induction. By equation \eqref{invariant_aggregation}, $W_0^{(\infty)}|_G \ne W_0^{(\infty)}|_H$ implies
    \begin{gather}
        \mathrm{AGGR}_0(\text{null};W_1^{(\infty)}(\cdot)|_G, \mathcal{R}_1(G))\ne \mathrm{AGGR}_0(\text{null};W_1^{(\infty)}(\cdot)|_H, \mathcal{R}_1(H)), 
    \end{gather}
    or simply
    \begin{gather}
        \mathcal{H}\left(\lbbr W_1^{(\infty)}(\tilde{\mathbf{u}})|_G:\tilde{\mathbf{u}}\in\mathcal{R}_1(G)\rbbr\right)\ne \mathcal{H}\left(\lbbr W_1^{(\infty)}(\tilde{\mathbf{v}})|_H:\tilde{\mathbf{v}}\in\mathcal{R}_1(H)\rbbr\right),\label{different_multiset_init}
    \end{gather}
    where $\mathcal{H}(\cdot)$ is the same injective hashing function for both sides. Equation \eqref{different_multiset_init} implies that either (i) $|\mathcal{R}_1(G)|\ne|\mathcal{R}_1(H)|$, or (ii) for any bijection $f:\mathcal{R}_1(G)\rightarrow \mathcal{R}_1(H)$, there exist $\tilde{\mathbf{u}}\in\mathcal{R}_1(G)$ and $\tilde{\mathbf{v}}=f(\tilde{\mathbf{u}})\in\mathcal{R}_1(H)$, such that 
    \begin{gather}
        W_1^{(\infty)}(\tilde{\mathbf{u}})|_G\ne W_1^{(\infty)}(\tilde{\mathbf{v}})|_H. \label{different_extra_init}
    \end{gather}
    If (i) is the case, Duplicator loses after the first putting move of the initialization round, since she fails to choose a bijection $f:\mathcal{R}_1(G)\rightarrow \mathcal{R}_1(H)$. Otherwise, for any bijection $f:\mathcal{R}_1(G)\rightarrow \mathcal{R}_1(H)$ chosen by Duplicator, Spoiler can put pebbles $p_1,\ldots, p_{i_1}$ at $\Delta\mathbf{u}_1=\tilde{\mathbf{u}}\in\mathcal{R}_1(G)$, thus forcing Duplicator to put pebbles $q_1,\ldots,q_{i_1}$ at $\Delta\mathbf{v}_1=\tilde{\mathbf{v}}\in\mathcal{R}_1(H)$; after the first putting move, we see that $\mathbf{u}_1=\Delta\mathbf{u}_1=\tilde{\mathbf{u}}$ and $\mathbf{v}_1=\Delta\mathbf{v}_1=\tilde{\mathbf{v}}$. Therefore, equation \eqref{different_extra_init} implies $W_1^{(\infty)}(\mathbf{u}_1)|_G\ne W_1^{(\infty)}(\mathbf{v}_1)|_H$. We have then proved Lemma \ref{theorem1_lemma1} for the case of $n=1$.

    Now assuming that Lemma \ref{theorem1_lemma1} holds for the case of $n=\tilde{n}-1$ with $2\leqslant \tilde{n}\leqslant N$, we now prove Lemma \ref{theorem1_lemma1} for the case of $n=\tilde{n}$. When $n=\tilde{n}$, it is supposed that Duplicator has not lost after the $(\tilde{n}-1)$-th putting move of the initialization round. By the inductive hypothesis, this implies that Spoiler can maintain 
    \begin{gather}
        W_{\tilde{n}-1}^{(\infty)}(\mathbf{u}_{\tilde{n}-1})|_G\ne W_{\tilde{n}-1}^{(\infty)}(\mathbf{v}_{\tilde{n}-1})|_H \label{different_colors_init_in}
    \end{gather}
    after the $(\tilde{n}-1)$-th putting move of the initialization round. By \eqref{invariant_aggregation}, equation \eqref{different_colors_init_in} means that
    \begin{gather}
        \mathrm{AGGR}_{i_{\tilde{n}-1}}(\mathbf{u}_{\tilde{n}-1}; W_{\tilde{n}}^{(\infty)}(\cdot)|_G,\mathcal{R}_{\tilde{n}}(G))\ne \mathrm{AGGR}_{i_{\tilde{n}-1}}(\mathbf{v}_{\tilde{n}-1}; W_{\tilde{n}}^{(\infty)}(\cdot)|_H,\mathcal{R}_{\tilde{n}}(H)),
    \end{gather}
    or equivalently,
    \begin{gather}
        \notag \mathcal{H}\left(\lbbr W_{\tilde{n}}^{(\infty)}((\mathbf{u}_{\tilde{n}-1},\tilde{\mathbf{u}}))|_G:\tilde{\mathbf{u}}\in\mathcal{R}_{\tilde{n}}(G)\backslash\mathbf{u}_{\tilde{n}-1}\rbbr\right)\\
        \ne \mathcal{H}\left(\lbbr W_{\tilde{n}}^{(\infty)}((\mathbf{v}_{\tilde{n}-1},\tilde{\mathbf{v}}))|_H:\tilde{\mathbf{v}}\in\mathcal{R}_{\tilde{n}}(H)\backslash\mathbf{v}_{\tilde{n}-1}\rbbr\right), \label{different_multiset_init_in}
    \end{gather}
    in which $\mathcal{H}(\cdot)$ is the same injective hashing function for both sides (but can be different from the one in equation \eqref{different_multiset_init}). Equation \eqref{different_multiset_init_in} implies either (i) $|\mathcal{R}_{\tilde{n}}(G)\backslash\mathbf{u}_{\tilde{n}-1}|\ne |\mathcal{R}_{\tilde{n}}(H)\backslash\mathbf{v}_{\tilde{n}-1}|$, or (ii) for any bijection $f:\mathcal{R}_{\tilde{n}}(G)\backslash\mathbf{u}_{\tilde{n}-1}\rightarrow\mathcal{R}_{\tilde{n}}(H)\backslash\mathbf{v}_{\tilde{n}-1}$, there exist $\tilde{\mathbf{u}}\in \mathcal{R}_{\tilde{n}}(G)\backslash\mathbf{u}_{\tilde{n}-1}$ and $\tilde{\mathbf{v}}=f(\tilde{\mathbf{u}})\in \mathcal{R}_{\tilde{n}}(H)\backslash\mathbf{v}_{\tilde{n}-1}$, such that
    \begin{gather}
        W_{\tilde{n}}^{(\infty)}((\mathbf{u}_{\tilde{n}-1},\tilde{\mathbf{u}}))|_G\ne W_{\tilde{n}}^{(\infty)}((\mathbf{v}_{\tilde{n}-1},\tilde{\mathbf{v}}))|_H.\label{different_extra_init_in}
    \end{gather}
    If (i) is the case, Duplicator loses after the $\tilde{n}$-th putting move of the initialization round. Otherwise, for any bijection $f:\mathcal{R}_{\tilde{n}}(G)\backslash\mathbf{u}_{\tilde{n}-1}\rightarrow\mathcal{R}_{\tilde{n}}(H)\backslash\mathbf{v}_{\tilde{n}-1}$ Duplicator chooses for the $\tilde{n}$-th putting move, Spoiler can put pebbles $p_{i_{\tilde{n}-1}+1},p_{i_{\tilde{n}-1}+2},\ldots, p_{i_{\tilde{n}}}$ at $\Delta\mathbf{u}_{\tilde{n}}=\tilde{\mathbf{u}}\in \mathcal{R}_{\tilde{n}}(G)\backslash\mathbf{u}_{\tilde{n}-1}$. Duplicator then puts $q_{i_{\tilde{n}-1}+1},q_{i_{\tilde{n}-1}+2},\ldots, q_{i_{\tilde{n}}}$ at $\Delta\mathbf{v}_{\tilde{n}}=\tilde{\mathbf{v}}\in \mathcal{R}_{\tilde{n}}(H)\backslash\mathbf{v}_{\tilde{n}-1}$. After the $\tilde{n}$-th putting move, we have $\mathbf{u}_{\tilde{n}}=(\mathbf{u}_{\tilde{n}-1},\tilde{\mathbf{u}})$ and $\mathbf{v}_{\tilde{n}}=(\mathbf{v}_{\tilde{n}-1},\tilde{\mathbf{v}})$. Therefore, equation \eqref{different_extra_init_in} implies $W_{\tilde{n}}^{(\infty)}(\mathbf{u}_{\tilde{n}})|_G\ne W_{\tilde{n}}^{(\infty)}(\mathbf{v}_{\tilde{n}})|_H$. We have thus proved Lemma \ref{theorem1_lemma1} for the case of $n=\tilde{n}$. We assert by induction that Lemma \ref{theorem1_lemma1} holds. \\\emph{End of proof of Lemma \ref{theorem1_lemma1}.}

    Now, using Lemma \ref{theorem1_lemma1}, we find that given $W(G)\ne W(H)$, either of the two cases happens: (i) Spoiler wins the game in the initialization round; (ii) Spoiler maintains
    \begin{gather}
        W^{(\infty)}(\mathbf{u}_N)\ne W^{(\infty)}(\mathbf{v}_N)\label{different_final_states}
    \end{gather}
    after the initialization round, where $\mathbf{u}_N\in\mathcal{R}(G)$ and $\mathbf{v}_N\in\mathcal{R}(H)$ are the $k$-tuples occupied by pebbles $p_1,\ldots, p_k$ and $q_1,\ldots, q_k$ respectively. Since the case (i) trivially implies \textbf{(b)}, it suffices to prove that (ii) leads to \textbf{(b)} in the following. 

    Equation \eqref{different_final_states} directly implies the following statement: $\exists T\geqslant 0, W^{(T)}(\mathbf{u}_N)\ne W^{(T)}(\mathbf{v}_N)$. We assert that
    \begin{lemma}\label{key_for_theorem1_forward}
        If $W^{(T)}(\mathbf{u}_N)\ne W^{(T)}(\mathbf{v}_N)$ with $T\geqslant 0$, then Spoiler wins the game no later than the $T$-th update round ends. (The initialization round is seen as the 0-th update round.)
    \end{lemma}

    If Lemma \ref{key_for_theorem1_forward} holds, we have already finished the proof of \textbf{(a)} $\Rightarrow$ \textbf{(b)}. To see why Lemma \ref{key_for_theorem1_forward} is true, we need another lemma whose proof we will omit because it resembles the proof of Lemma \ref{theorem1_lemma1}:
    \begin{lemma}\label{theorem1_lemma2}
        Assume that $W^{(T)}(\mathbf{u}_N)\ne W^{(T)}(\mathbf{v}_N)$ with some $T\geqslant 1$, in which $\mathbf{u}_N$ and $\mathbf{v}_N$ are the $k$-tuples occupied by the main pebbles in $G$ and $H$ after the initialization round, respectively. The following statement holds for all $m=1,2,\ldots, M$: if Duplicator has not lost after the $(m-1)$-th putting move of the first update round, then either (i) Duplicator loses after the $m$-th putting move of the first update round, or (ii) Spoiler can maintain that
        \begin{gather}
            \mathrm{MSG}_m^{(T-1)}(\mathbf{u}'_m;\mathbf{u}_N)|_G\ne \mathrm{MSG}_m^{(T-1)}(\mathbf{v}'_m;\mathbf{v}_N)|_H,
        \end{gather}
        after the $m$-th putting move of the first update round.
    \end{lemma}
    Lemma \ref{theorem1_lemma2} is a complete analogue to Lemma \ref{theorem1_lemma1}. Now we can prove Lemma \ref{key_for_theorem1_forward}.

    \emph{Proof of Lemma \ref{key_for_theorem1_forward}.} We prove by induction. For the case of $T=0$, $W^{(0)}(\mathbf{u}_N)\ne W^{(0)}(\mathbf{v}_N)$ means that 
    \begin{gather}
        \mathrm{atp}(\mathbf{u}_N)\ne\mathrm{atp}(\mathbf{v}_N), \label{different_isotype_at_init}
    \end{gather}
    because $W^{(0)}(\cdot)=\Phi(\mathrm{atp}(\cdot))$ is a function of $\mathrm{atp}(\cdot)$. Equation \eqref{different_isotype_at_init} immediately implies that Spoiler wins after the initialization round. We have thus proved the case of $T=0$.

    Now, assume that Lemma \ref{key_for_theorem1_forward} holds for the case of $T=\tilde{T}-1$, in which $\tilde{T}\geqslant 1$. We have to prove that Lemma \ref{key_for_theorem1_forward} also holds for $T=\tilde{T}$. When $T=\tilde{T}$, it is assumed that $W^{(\tilde{T})}(\mathbf{u}_N)\ne W^{(\tilde{T})}(\mathbf{v}_N)$. Since $\tilde{T}\geqslant 1$, we can make use of Lemma \ref{theorem1_lemma2} (taking $m=M$) to see that either of the two cases happens: (i) Duplicator loses in one of the $M$ putting moves of the first update round; (ii) after all $M$ putting moves, Spoiler can maintain
    \begin{gather}
        \mathrm{MSG}^{(\tilde{T}-1)}(\mathbf{u}'_M;\mathbf{u}_N)|_G\ne\mathrm{MSG}^{(\tilde{T}-1)}(\mathbf{v}'_M;\mathbf{v}_N)|_H,
    \end{gather}
    in which $\mathbf{u}'_M$ and $\mathbf{v}'_M$ are the $t$-tuples occupied by pebbles $p_{k+1},p_{k+2},\ldots, p_{k+t}$ and $q_{k+1},q_{k+2},\ldots, q_{k+t}$ respectively. If (i) is the case, we immediately know that Lemma \ref{key_for_theorem1_forward} is true for $T=\tilde{T}$. Otherwise, by equation \eqref{message_definition} we know that either
    \begin{gather}
        \Psi(\mathrm{atp}((\mathbf{u}_N,\mathbf{u}'_M)))\ne \Psi(\mathrm{atp}((\mathbf{v}_N,\mathbf{v}'_M))),
    \end{gather}
    or $\exists c\in[C]$, such that
    \begin{gather}
        W^{(\tilde{T}-1)}\left((\mathbf{u}_N)_c(\mathbf{u}'_M)\right)|_G\ne W^{(\tilde{T}-1)}\left((\mathbf{v}_N)_c(\mathbf{v}'_M)\right)|_H. \label{different_substitutes}
    \end{gather}
    If the former is the case, we know that Spoiler already wins in the first update round, since $(\mathbf{u}_N,\mathbf{u}'_M)$ and $(\mathbf{v}_N,\mathbf{v}'_M)$ have different isomorphism types. This again leads to Lemma \ref{key_for_theorem1_forward} with $T=\tilde{T}\geqslant 1$. If the latter is the case, let us notice that in equation \eqref{different_substitutes}, $(\mathbf{u}_N)_c(\mathbf{u}'_M)=\mathrm{REPLACE}(\mathbf{u}_N, \mathbf{u}'_M;r_1^{(c)},\ldots r_k^{(c)})$ is exactly the $k$-tuple occupied by the pebbles $p_{r_1^{(c)}},\ldots, p_{r_k^{(c)}}$ after all $M$ putting moves; similarly, $(\mathbf{v}_N)_c(\mathbf{v}'_M)$ is the $k$-tuple occupied by pebbles $q_{r_1^{(c)}},\ldots, q_{r_k^{(c)}}$ after the $M$ putting moves. From Spoiler's perspective, he finds that by choosing the sequence $r_1^{(c)},\ldots, r_k^{(c)}$ in the subsequent removing move, he can ensure that the pebbles $p_1,\ldots, p_k$ and $q_1,\ldots, q_k$ should be moved to $(\mathbf{u}_N)_c(\mathbf{u}'_M)$ and $(\mathbf{v}_N)_c(\mathbf{v}'_M)$, respectively. By equation \eqref{different_substitutes}, Spoiler can actually ensure that
    \begin{gather}
        W^{(\tilde{T}-1)}(\mathbf{u}^*_N)|_G\ne W^{(\tilde{T}-1)}(\mathbf{v}^*_N)|_H,\label{induction_to_next_lemma2}
    \end{gather}
    in which $\mathbf{u}^*_N$ and $\mathbf{v}^*_N$ are the $k$-tuples occupied by pebbles $p_1,\ldots, p_k$ and $q_1,\ldots, q_k$ when the first update round ends, respectively. By the inductive hypothesis, equation \eqref{induction_to_next_lemma2} implies that Spoiler wins the game in at most another $(\tilde{T}-1)$ rounds. Therefore, Spoiler wins no later than the $\tilde{T}$-th update round ends. We assert by induction that Lemma \ref{key_for_theorem1_forward} is true. \\\emph{End of proof of Lemma \ref{key_for_theorem1_forward}.}

    Our desired result is now straightforward.

    \textbf{$\bm{\neg}$(a) $\bm{\Rightarrow}$ $\bm{\neg}$(b).} Since $W(G)=W(H)$, we have $W_0^{(\infty)}|_G=W_0^{(\infty)}|_H$. In correspondence to Lemma \ref{theorem1_lemma1}, we have

    \begin{lemma}\label{theorem1_lemma3}
        If $W_0^{(\infty)}|_G=W_0^{(\infty)}|_H$, then Duplicator can maintain that 
        \begin{gather}
            W_n^{(\infty)}(\mathbf{u}_n)|_G=W_n^{(\infty)}(\mathbf{v}_n)|_H
        \end{gather}
        after the $n$-th putting move of the initialization round, for all $n=1,2,\ldots, N$.
    \end{lemma}
    \emph{Proof of Lemma \ref{theorem1_lemma3}.} We prove by induction. The assumption $W_0^{(\infty)}|_G=W_0^{(\infty)}|_H$ means that equation \eqref{different_multiset_init} is false, or equivalently,
    \begin{gather}
        \lbbr W_1^{(\infty)}(\tilde{\mathbf{u}})|_G:\tilde{\mathbf{u}}\in\mathcal{R}_1(G)\rbbr=\lbbr W_1^{(\infty)}(\tilde{\mathbf{v}})|_H:\tilde{\mathbf{v}}\in\mathcal{R}_1(H)\rbbr.
    \end{gather}
    In other words, there exists a bijection $f:\mathcal{R}_1(G)\rightarrow \mathcal{R}_1(H)$ such that 
    \begin{gather}
        \forall \tilde{\mathbf{u}}\in\mathcal{R}_1(G), \quad W_1^{(\infty)}(\tilde{\mathbf{u}})|_G=W_1^{(\infty)}(f(\tilde{\mathbf{u}}))|_H. \label{condition_for_init}
    \end{gather}
    By choosing the bijection $f$ that satisfies \eqref{condition_for_init}, Duplicator can maintain that $W_1^{(\infty)}(\mathbf{u}_1)|_G=W_1^{(\infty)}(\mathbf{v}_1)|_H$ after the first putting move of the initialization round. We have thus proved the case of $n=1$.

    If Lemma \ref{theorem1_lemma3} holds for $n=\tilde{n}-1$ with some $2\leqslant \tilde{n}\leqslant N$, we know that Duplicator can already maintain that 
    \begin{gather}
        W_{\tilde{n}-1}^{(\infty)}(\mathbf{u}_{\tilde{n}-1})|_G=W_{\tilde{n}-1}^{(\infty)}(\mathbf{v}_{\tilde{n}-1})|_H\label{assumption_for_next_init}
    \end{gather}
    after the $(\tilde{n}-1)$-th putting move of the initialization round. Equation \eqref{assumption_for_next_init} means that \eqref{different_multiset_init_in} is false, which implies that there exists a bijection $f:\mathcal{R}_{\tilde{n}}(G)\backslash\mathbf{u}_{\tilde{n}-1}\rightarrow\mathcal{R}_{\tilde{n}}(H)\backslash\mathbf{v}_{\tilde{n}-1}$, such that
    \begin{gather}
        \forall \tilde{\mathbf{u}}\in \mathcal{R}_{\tilde{n}}(G)\backslash\mathbf{u}_{\tilde{n}-1}, \quad W_{\tilde{n}}^{(\infty)}((\mathbf{u}_{\tilde{n}-1},\tilde{\mathbf{u}}))|_G=W_{\tilde{n}}^{(\infty)}((\mathbf{v}_{\tilde{n}-1},f(\tilde{\mathbf{u}})))|_H.\label{condition_for_step}
    \end{gather}
    Choosing the bijection $f$ that satisfies \eqref{condition_for_step} enables Duplicator to maintain $W_{\tilde{n}}^{(\infty)}(\mathbf{u}_{\tilde{n}})|_G=W_{\tilde{n}}^{(\infty)}(\mathbf{v}_{\tilde{n}})|_H$ after the next putting move (i.e., the $\tilde{n}$-th putting move of the initialization round). We have thus made the induction step. Now we assert that Lemma \ref{theorem1_lemma3} holds. \\\emph{End of proof of Lemma \ref{theorem1_lemma3}.}

    To get our desired result, we still need the following lemma:

    \begin{lemma}\label{key_for_theorem1_backward}
        Assume that $W^{(T)}(\mathbf{u}_N)|_G=W^{(T)}(\mathbf{v}_N)|_H$ with some $T\geqslant 1$, in which $\mathbf{u}_N$ and $\mathbf{v}_N$ are the $k$-tuples occupied by the main pebbles in $G$ and $H$ after the initialization round, respectively. Then, Duplicator can maintain $W^{(T-1)}(\mathbf{u}_N^*)|_G=W^{(T-1)}(\mathbf{v}_N^*)|_H$ after the first update round, in which $\mathbf{u}_N^*$ and $\mathbf{v}_N^*$ are the $k$-tuples occupied by the main pebbles in $G$ and $H$ after the first update round, respectively.
    \end{lemma}

    To prove Lemma \ref{key_for_theorem1_backward}, we make use of the following Lemma \ref{theorem1_lemma4}. Due to its similarity to Lemma \ref{theorem1_lemma3}, we omit its proof.
    \begin{lemma}\label{theorem1_lemma4}
        If $W^{(T)}(\mathbf{u}_N)|_G=W^{(T)}(\mathbf{v}_N)|_H$ with some $T\geqslant 1$, then Duplicator can maintain that 
        \begin{gather}
            \mathrm{MSG}_m^{(T-1)}(\mathbf{u}'_m;\mathbf{u}_N)|_G=\mathrm{MSG}_m^{(T-1)}(\mathbf{v}'_m;\mathbf{v}_N)|_H
        \end{gather}
        after the $m$-th putting move of the first update round, for all $m=1,2,\ldots, M$.
    \end{lemma}
    Taking $m=M$ in Lemma \ref{theorem1_lemma4}, we find that $W^{(T)}(\mathbf{u}_N)|_G=W^{(T)}(\mathbf{v}_N)|_H$ implies that Duplicator can maintain
    \begin{gather}
        \mathrm{MSG}^{(T-1)}(\mathbf{u}'_M;\mathbf{u}_N)|_G=\mathrm{MSG}^{(T-1)}(\mathbf{v}'_M;\mathbf{v}_N)|_H
    \end{gather}
    after all $M$ putting moves of the first update round. By \eqref{message_definition}, this means that $\mathrm{atp}((\mathbf{u}_N,\mathbf{u}'_M))=\mathrm{atp}((\mathbf{v}_N,\mathbf{v}'_M))$ and that $\forall c\in[C]$, $W^{(T-1)}\left((\mathbf{u}_N)_c(\mathbf{u}'_M)\right)|_G=W^{(T-1)}\left((\mathbf{v}_N)_c(\mathbf{v}'_M)\right)|_H$. The former condition guarantees that Duplicator has not lost after $M$ putting moves, while the latter means that no matter what sequence $r_1^{(c)},\ldots, r_k^{(c)}$ Spoiler chooses, the removing move must end up with $p_1,\ldots, p_k$ and $q_1,\ldots, q_k$ moved to some $\mathbf{u}_N^*$ and $\mathbf{v}_N^*$ that satisfy $W^{(T-1)}(\mathbf{u}_N^*)|_G=W^{(T-1)}(\mathbf{v}_N^*)|_H$. Lemma \ref{key_for_theorem1_backward} is thus proved.

    Now we may turn to proving $\neg$\textbf{(a)} $\Rightarrow$ $\neg$\textbf{(b)}. Taking $n=N$ in Lemma \ref{theorem1_lemma3}, we find that given $W(G)=W(H)$, Duplicator can maintain that 
    \begin{gather}
        W^{(\infty)}(\mathbf{u}_N)|_G=W^{(\infty)}(\mathbf{v}_N)|_H\label{final_same}
    \end{gather}
    at the end of the initialization round, where $\mathbf{u}_N$ and $\mathbf{v}_N$ are the $k$-tuples occupied by pebbles $p_1,\ldots, p_k$ and $q_1,\ldots, q_k$, respectively. Since $W^{(\infty)}(\cdot)|_G$ and $W^{(\infty)}(\cdot)|_H$ are stable colorings of $k$-tuples, there exists $T^*\geqslant 1$ such that
    \begin{gather}
        \forall \mathbf{u}\in\mathcal{R}(G),\quad W^{(T^*)}(\mathbf{u})|_G=W^{(T^*-1)}(\mathbf{u})|_G=W^{(\infty)}(\mathbf{u})|_G,\label{stable_G}\\
        \forall \mathbf{v}\in\mathcal{R}(H),\quad W^{(T^*)}(\mathbf{v})|_H=W^{(T^*-1)}(\mathbf{v})|_H=W^{(\infty)}(\mathbf{v})|_H.\label{stable_H}
    \end{gather}
    Therefore, equation \eqref{final_same} implies that $W^{(T^*)}(\mathbf{u}_N)|_G=W^{(T^*)}(\mathbf{v}_N)|_H$. By Lemma \ref{key_for_theorem1_backward}, Duplicator can maintain $W^{(T^*-1)}(\mathbf{u}_N^*)|_G=W^{(T^*-1)}(\mathbf{v}_N^*)|_H$, which by \eqref{stable_G} and \eqref{stable_H} is equivalent to $W^{(T^*)}(\mathbf{u}_N^*)|_G=W^{(T^*)}(\mathbf{v}_N^*)|_H$---a result that allows us to apply the above discussion repeatedly. We eventually find that Duplicator never loses the game.
\end{proof}

\section{Proof of Theorem~\ref{theorem:cops-robber-ehren-frai}}\label{app:proof2}

\begin{proof}
    \textbf{(a) $\bm{\Rightarrow}$ (b).} The key idea of our proof is to first construct \emph{a set of} graphs $\mathcal{S}\subseteq\mathcal{G}$, on each of which Cops win the Cops-Robber game, and then try to prove that \emph{at least one} graph $F\in\mathcal{S}$ satisfies $\mathrm{Hom}(F,G)\ne\mathrm{Hom}(F,H)$, given that \textbf{(a)} is true.
    
    We introduce the following notation: given $G\in\mathcal{G}$ and $\mathbf{u}=(u_1,\ldots, u_\ell)\in\mathcal{V}_G^\ell$, let $G[\mathbf{u}]$ be the subgraph of $G$ induced by $\{u_1,\ldots, u_\ell\}$. It is easy to see that $G[\mathbf{u}]$ is uniquely determined by $\mathrm{atp}(\mathbf{u})$, up to isomorphism. 

    \begin{definition}[Spoiler's record graph]\label{record_graph}
        Let $G\in\mathcal{G}$. Assume that Spoiler and Duplicator are playing the Ehrenfeucht-Fra\"iss\'e game on two copies of $G$. For every putting move, Duplicator always chooses the identity mapping as $f$. (Therefore, Spoiler never wins the game after any number of rounds.) Spoiler's actions in the above game are fully determined by the following sequence,
        \begin{align}
            \notag a=\Big(\Delta\mathbf{u}_1^{(0)},\ldots, \Delta\mathbf{u}_N^{(0)},&\Delta\mathbf{u}'^{(1)}_1,\ldots, \Delta\mathbf{u}'^{(1)}_M, (r_1^{(c_1)},\ldots, r_k^{(c_1)}),\\
            &\Delta\mathbf{u}'^{(2)}_1,\ldots, \Delta\mathbf{u}'^{(2)}_M, (r_1^{(c_2)},\ldots, r_k^{(c_2)}),\ldots,\ldots\Big),
        \end{align}
        in which $\Delta\mathbf{u}_n^{(0)}$ ($1\leqslant n\leqslant N$) is the tuple chosen by Spoiler in the $n$-th putting move of the initialization round, $\Delta\mathbf{u}'^{(T)}_m$ ($T\geqslant 1, 1\leqslant m\leqslant M$) is the tuple chosen by Spoiler in the $m$-th putting move of the $T$-th update round, and $(r_1^{(c_T)},\ldots, r_k^{(c_T)})$ ($T\geqslant 1$) is the sequence chosen by Spoiler in the removing move of the $T$-th update round. Based on the action sequence $a$, we define a series of \textbf{Spoiler's record graphs} as following:
        \begin{itemize}
            \item Construct $F^{(0)}_n(G,a)$ ($n=0,1,\ldots, N$) via the following steps:
            \begin{itemize}
                \item[$\circ$] Let $F^{(0)}_0(G,a)$ be an empty graph (whose node set and edge set are both $\varnothing$). For the first move of the initialization round, Spoiler puts pebbles $p_1,\ldots, p_{i_1}$ at $\mathbf{u}_1^{(0)}=\Delta\mathbf{u}_1^{(0)}=(u_1^{(0)},\ldots, u_{i_1}^{(0)})\in\mathcal{R}_1(G)$. Let $\tilde{i}_1$ be the number of nodes occupied by pebbles $p_1,\ldots, p_{i_1}$ ($\tilde{i}_1$ can be smaller than $i_1$, since two pebbles can occupy the same node). The graph $F^{(0)}_1(G,a)$ is obtained by first adding $\tilde{i}_1$ nodes to $F^{(0)}_0(G,a)$, and then adding a minimum number of edges such that we can find $w_1^{(0)},\ldots, w_{i_1}^{(0)}\in\mathcal{V}_{F^{(0)}_1(G,a)}$ that makes
                \begin{gather}
                    \varphi_1^{(0)}:\quad u_1^{(0)}\mapsto w_1^{(0)},\cdots, u_{i_1}^{(0)}\mapsto w_{i_1}^{(0)}
                \end{gather}
                an isomorphism from $G[\mathbf{u}_1^{(0)}]$ to $F^{(0)}_1(G,a)$. 
                \item[$\circ$] For the $n$-th ($2\leqslant n\leqslant N$) putting move of the initialization round, Spoiler puts pebbles $p_{i_{n-1}+1},\ldots, p_{i_n}$ at $\Delta\mathbf{u}_n^{(0)}=(u_{i_{n-1}+1}^{(0)},\ldots, u_{i_n}^{(0)})\in\mathcal{R}_n(G)\backslash\mathbf{u}^{(0)}_{n-1}$. Let $\tilde{i}_n$ be the number of nodes occupied by pebbles $p_{i_{n-1}+1},\ldots, p_{i_n}$. Notice that there is already an isomorphism
                \begin{gather}
                    \varphi_{n-1}^{(0)}:\quad u_1^{(0)}\mapsto w_1^{(0)},\cdots, u_{i_{n-1}}^{(0)}\mapsto w_{i_{n-1}}^{(0)}
                \end{gather}
                from $G[\mathbf{u}_{n-1}^{(0)}]$ to $F^{(0)}_{n-1}(G,a)$. The graph $F^{(0)}_n(G,a)$ is obtained by first adding $\tilde{i}_n$ nodes to $F^{(0)}_{n-1}(G,a)$, and then adding a minimum number of edges such that by extending $\varphi_{n-1}^{(0)}$ to
                \begin{gather}
                    \varphi_n^{(0)}=\varphi_{n-1}^{(0)}\cup \left\{u_{i_{n-1}+1}^{(0)}\mapsto w_{i_{n-1}+1}^{(0)},\cdots, u_{i_n}^{(0)}\mapsto w_{i_n}^{(0)}\right\},
                \end{gather}
                with some $w_{i_{n-1}+1}^{(0)},\ldots, w_{i_n}^{(0)}\in\mathcal{V}_{F^{(0)}_n(G,a)}-\mathcal{V}_{F^{(0)}_{n-1}(G,a)}$, we can make $\varphi_n^{(0)}$ an isomorphism from $G[\mathbf{u}_n^{(0)}]$ to $F^{(0)}_n(G,a)$, where $\mathbf{u}_n^{(0)}=(\mathbf{u}_{n-1}^{(0)},\Delta\mathbf{u}_n^{(0)})$ as before.
                \item[$\circ$] Let $F^{(0)}(G,a)$ be an alias of $F^{(0)}_N(G,a)$. Also let
                \begin{gather}
                    \varphi^{(0)}=\varphi_N^{(0)}:\quad u_1^{(0)}\mapsto w_1^{(0)},\cdots, u_k^{(0)}\mapsto w_k^{(0)},
                \end{gather}
                which is an isomorphism from $G[\mathbf{u}_N^{(0)}]$ to $F^{(0)}(G,a)$, in which $\mathbf{u}_N^{(0)}=(u_1^{(0)},\ldots, u_k^{(0)})$ is the $k$-tuple occupied by pebbles $p_1,\ldots, p_k$ after the initialization round.
            \end{itemize}
            \item For each of $T=1,2,\ldots$, construct $F^{(T)}_m(G,a)$ ($m=0,1,\ldots, M$) via the following steps:
            \begin{itemize}
                \item[$\circ$] Start with $F^{(T)}_0(G,a)=F^{(T-1)}(G,a)$ and
                \begin{gather}
                    \varphi^{(T)}_0=\varphi^{(T-1)}:\quad u_1^{(T-1)}\mapsto w_1^{(T-1)},\cdots, u_k^{(T-1)}\mapsto w_k^{(T-1)},
                \end{gather}
                in which $\mathbf{u}^{(T-1)}_N=(u_1^{(T-1)},\ldots, u_k^{(T-1)})$ is the $k$-tuple occupied by the main pebbles after the $(T-1)$-th update round, and $w_1^{(T-1)},\ldots, w_k^{(T-1)}\in\mathcal{V}_{F^{(T-1)}(G,a)}$. For the $m$-th ($1\leqslant m\leqslant M$) putting move of the $T$-th update round, Spoiler puts pebbles $p_{k+j_{m-1}+1},\ldots, p_{k+j_m}$ on $\Delta\mathbf{u}'^{(T)}_m=(u'^{(T)}_{j_{m-1}+1},\ldots, u'^{(T)}_{j_m})\in\mathcal{F}_m(G,\mathbf{u}_N^{(T-1)})\backslash \mathbf{u}'^{(T)}_{m-1}$. Denote as $\tilde{j}_m$ the number of nodes occupied by pebbles $p_{k+j_{m-1}+1},\ldots, p_{k+j_m}$. We construct $F^{(T)}_m(G,a)$ by first adding $\tilde{j}_m$ nodes to $F^{(T)}_{m-1}(G,a)$, and then adding a minimum number of edges such that $\varphi^{(T)}_{m-1}$ can be extended to
                \begin{gather}
                    \varphi^{(T)}_m=\varphi^{(T)}_{m-1}\cup \left\{u'^{(T)}_{j_{m-1}+1}\mapsto w^{(T-1)}_{k+j_{m-1}+1},\cdots, u'^{(T)}_{j_m}\mapsto w^{(T-1)}_{k+j_m}\right\},
                \end{gather}
                with some $w^{(T-1)}_{k+j_{m-1}+1},\ldots, w^{(T-1)}_{k+j_m}\in\mathcal{V}_{F^{(T)}_m(G,a)}-\mathcal{V}_{F^{(T)}_{m-1}(G,a)}$, and that $\varphi^{(T)}_m$ is an isomorphism from $G[(\mathbf{u}_N^{(T-1)},\mathbf{u}'^{(T)}_m)]$ to the subgraph of $F^{(T)}_m(G,a)$ induced by $(w_1^{(T-1)},\ldots, w^{(T-1)}_{k+j_m})$, where $\mathbf{u}'^{(T)}_m=(\mathbf{u}'^{(T)}_{m-1},\Delta \mathbf{u}'^{(T)}_m)$ as before. Finally, let $F^{(T)}(G,a)$ be an alias of $F^{(T)}_M(G,a)$. 
                \item[$\circ$] For the removing move of the $T$-th update round, Spoiler chooses the sequence $(r_1^{(c_T)},\ldots, r_k^{(c_T)})$. After the removing move, let $\mathbf{u}_N^{(T)}=(u_1^{(T)},\ldots, u_k^{(T)})$ be the new $k$-tuple occupied by pebbles $p_1,\ldots, p_k$. We then let $\varphi^{(T)}$ be 
                \begin{gather}
                    \varphi^{(T)}:\quad u_1^{(T)}\mapsto w_1^{(T)}=w^{(T-1)}_{r_1^{(c_T)}},\ldots, u^{(T)}_k\mapsto w_k^{(T)}=w^{(T-1)}_{r_k^{(c_T)}}.
                \end{gather}
                One can easily show that $\varphi^{(T)}$ is an isomorphism from $G[\mathbf{u}_N^{(T)}]$ to the subgraph of $F^{(T)}(G,a)$ induced by $(w_1^{(T)},\ldots, w_k^{(T)})$.
            \end{itemize}
        \end{itemize}
        The graph $F^{(T)}_p(G,a)$ is called Spoiler's record graph up to the $p$-th move of the $T$-th round, where $T\geqslant 0$, and $p$ is an integer in $\{0,1,\ldots, N\}$ for $T=0$, or in $\{0,1,\ldots, M\}$ for $T\geqslant 1$.
    \end{definition}

    The following lemma is straightforward.

    \begin{lemma}\label{induced_subgraph_rel}
        Let $\preccurlyeq$ be a partial order on $\mathcal{G}$ such that $\forall G,H\in\mathcal{G}$, $G\preccurlyeq H$ if and only if $G$ is an induced subgraph of $H$. Then we have 
        \begin{align}
            \notag F^{(0)}_0(G,a)\preccurlyeq F^{(0)}_1(G,a)\preccurlyeq\cdots&\preccurlyeq F^{(0)}_N(G,a)= F^{(1)}_0(G,a)\preccurlyeq F^{(1)}_1(G,a)\preccurlyeq \cdots\\
            &\preccurlyeq F^{(1)}_M(G,a)=F^{(2)}_0(G,a)\preccurlyeq F^{(2)}_1(G,a)\preccurlyeq\cdots\preccurlyeq\cdots,
        \end{align}
        for any $G\in\mathcal{G}$ and any action sequence $a$ of Spoiler on $G$.
    \end{lemma}

    The following lemma is essential for our proof.

    \begin{lemma}\label{cops_robber_win_record}
        Given any $T\geqslant 0$ and $p\in[N]$ (if $T=0$) or $p\in[M]$ (if $T\geqslant 1$), Cops win the Cops-Robber game on $F^{(T)}_p(G,a)$, with any $G\in\mathcal{G}$ and any action sequence $a$ of Spoiler on $G$.
    \end{lemma}
    \emph{Proof of Lemma \ref{cops_robber_win_record}.} Cops' winning strategy is given below: for the $n$-th ($1\leqslant n\leqslant N$) guarding move of the initialization round, they put new pebbles at $w_{i_{n-1}+1}^{(0)},\ldots, w_{i_n}^{(0)}$; for the $m$-th ($1\leqslant m\leqslant M$) guarding move of the $\tilde{T}$-th ($\tilde{T}\geqslant 1$) update round, they put new pebbles at $w_{k+j_{m-1}+1}^{(\tilde{T}-1)},\ldots, w_{k+j_m}^{(\tilde{T}-1)}$; for the unguarding move of the $\tilde{T}$-th ($\tilde{T}\geqslant 1$) update round, they choose the sequence $(r_1^{(c_{\tilde{T}})},\ldots, r_k^{(c_{\tilde{T}})})$. 

    One can verify that by implementing the above strategy, Cops can guarantee that the set of edges $C$ chosen by Robber after the $\tilde{p}$-th guarding move of the $\tilde{T}$-th update round ($1\leqslant \tilde{p}\leqslant N$ for $\tilde{T}=0$ and $1\leqslant \tilde{p}\leqslant M$ for $\tilde{T}\geqslant 1$) satisfies $C\cap\mathcal{E}_{F^{(\tilde{T})}_{\tilde{p}}(G,a)}=\varnothing$; additionally, $C$ remains unchanged after every unguarding move. Therefore, after the $p$-th guarding move of the $T$-th update round, Robber must lose. \\\emph{End of proof of Lemma \ref{cops_robber_win_record}.}

    Now we can give the construction of $\mathcal{S}$. Given $T\geqslant 0$ and $p\in[N]$ (for $T=0$) or $p\in[M]$ (for $T\geqslant 1$), define
    \begin{gather}
        \mathcal{S}^{(T)}_p=\{F^{(T)}_p(G,a):G\in\mathcal{G},a\text{ is an action sequence of Spoiler on }G\}.
    \end{gather}
    Due to \textbf{(a)}, there exists $T^*\geqslant 0$ such that Spoiler can guarantee to win the Ehrenfeucht-Fra\"iss\'e game no later than the $T^*$-th update round ends. We then let
    \begin{gather}
        \mathcal{S}=\bigcup_{T=0}^{T^*}\bigcup_{p=1}^{\substack{N\text{, if }T=0\\M\text{, if }T\geqslant 1}} \mathcal{S}^{(T)}_p.\label{definition_S}
    \end{gather}

    \begin{definition}
        Given $G\in\mathcal{G}$ and an action sequence $a$ of Spoiler on $G$, each of the following tuples is called a \textbf{bag}:
        \begin{align}
            &\mathbf{w}_n^{(0)}=(w_1^{(0)},\ldots, w_{i_n}^{(0)})\in \mathcal{V}_{F^{(0)}_n(G,a)}^{i_n}, & &1\leqslant n\leqslant N,\\
            &\mathbf{w}_m^{(T)}=(w_1^{(T-1)},\ldots, w_{k+j_m}^{(T-1)})\in \mathcal{V}_{F^{(T)}_m(G,a)}^{k+j_m}, & &T\geqslant 1, 1\leqslant m\leqslant M.
        \end{align}
        They are exactly the ranges of $\varphi_n^{(0)}$ (for $\mathbf{w}_n^{(0)}$) or $\varphi_m^{(T)}$ (for $\mathbf{w}_m^{(T)}$).
    \end{definition}

    Due to Lemma \ref{induced_subgraph_rel}, $F^{(T)}_p(G,a)$ contains all bags $\mathbf{w}^{(\tilde{T})}_{\tilde{p}}$ that satisfy $\tilde{T}<T$ or $\tilde{T}=T,\tilde{p}\leqslant p$.

    \begin{lemma}\label{finiteness_S}
        $\mathcal{S}$ is a finite set.
    \end{lemma}
    \emph{Proof of Lemma \ref{finiteness_S}.} For given $T$ and $p$ values, the number of bags present in $F^{(T)}_p(G,a)$ is determined. Furthermore, $F^{(T)}_p(G,a)$ is uniquely determined up to isomorphism by the isomorphism types of the bags it contains. Since the number of possible isomorphism types of each bag is finite, the set $\mathcal{S}^{(T)}_p$ is finite, given any $T\geqslant 0$, and $p\in[N]$ for $T=0$ or $p\in[M]$ for $T\geqslant 1$. Hence $\mathcal{S}$ is finite.\\\emph{End of proof of Lemma \ref{finiteness_S}.}

    \begin{definition}
        Let $F^{(T)}_p(G,a)$ be a Spoiler's record graph. For any $H\in\mathcal{G}$, a \textbf{bag-wise isomorphic homomorphism} from $F^{(T)}_p(G,a)$ to $H$ is a homomorphism $h:\mathcal{V}_{F^{(T)}_p(G,a)}\rightarrow\mathcal{V}_H$, such that for every bag $\mathbf{w}$ present in $F^{(T)}_p(G,a)$, $h$ is an isomorphism from $F^{(T)}_p(G,a)[\mathbf{w}]$ to $H[h(\mathbf{w})]$. The number of bag-wise isomorphic homomorphisms from $F^{(T)}_p(G,a)$ to $H$ is denoted as $\mathrm{bIsoHom}(F^{(T)}_p(G,a),H)$.
    \end{definition}

    \begin{lemma}\label{different_bisohom}
        Assume that Spoiler wins the Ehrenfeucht-Fra\"iss\'e game on $G$ and $H$ no later than the $T^*$-th update round ends. Then $\exists F\in\mathcal{S}$ that satisfies
        \begin{gather}
            \mathrm{bIsoHom}(F,G)\ne\mathrm{bIsoHom}(F,H).
        \end{gather}
    \end{lemma}
    \emph{Proof of Lemma \ref{different_bisohom}.} Assume that the contrary is the case. Namely, $\forall F\in\mathcal{S}$, $\mathrm{bIsoHom}(F,G)=\mathrm{bIsoHom}(F,H)$. We can prove that Duplicator will not lose the game in the first $(T^*+1)$ rounds, by explicitly constructing a strategy for Duplicator.

    By our assumption above, given any $F^{(0)}_1(\tilde{G},\tilde{a})\in\mathcal{S}_1^{(0)}$, we have $\mathrm{bIsoHom}(F^{(0)}_1(\tilde{G},\tilde{a}),G)=\mathrm{bIsoHom}(F^{(0)}_1(\tilde{G},\tilde{a}),H)$, where $\tilde{G}\in\mathcal{G}$ is an arbitrary graph, while $\tilde{a}$ is an action sequence on $\tilde{G}$. Since $F^{(0)}_1(\tilde{G},\tilde{a})$ contains only one bag $\mathbf{w}^{(0)}_1$, the above fact implies
    \begin{gather}
        \big|\{\mathbf{u}_1^{(0)}\in\mathcal{V}_G^{i_1}:\mathrm{atp}(\mathbf{u}_1^{(0)})=\mathrm{atp}(\mathbf{w}_1^{(0)})\}\big|=\big|\{\mathbf{v}_1^{(0)}\in\mathcal{V}_H^{i_1}:\mathrm{atp}(\mathbf{v}_1^{(0)})=\mathrm{atp}(\mathbf{w}_1^{(0)})\}\big|.\label{atp_same_10}
    \end{gather}
    Equation \eqref{atp_same_10} holds for any $F^{(0)}_1(\tilde{G},\tilde{a})\in\mathcal{S}_1^{(0)}$, in which different choices of $\tilde{G}$ and $\tilde{a}$ may lead to different $\mathrm{atp}(\mathbf{w}_1^{(0)})$. If $\mathrm{atp}(\mathbf{w}_1^{(0)})$ can traverse over all possible isomorphism types that a $\mathbf{u}_1^{(0)}\in\mathcal{R}_1(G)$ or $\mathbf{v}_1^{(0)}\in\mathcal{R}_1(H)$ can possess, we can conclude that Duplicator can choose a bijection $f: \mathcal{R}_1(G)\rightarrow \mathcal{R}_1(H)$ and ensure that $\mathrm{atp}(\mathbf{u}_1^{(0)})=\mathrm{atp}(\mathbf{v}_1^{(0)})$ for any $\mathbf{u}_1^{(0)}\in\mathcal{R}_1(G)$ and $\mathbf{v}_1^{(0)}=f(\mathbf{u}_1^{(0)})\in\mathcal{R}_1(H)$. In other words, Duplicator will not lose after the first putting move of the initialization round. 

    We now show that this is indeed the case; namely, by choosing different $\tilde{G}$ and $\tilde{a}$, the corresponding $\mathrm{atp}(\mathbf{w}_1^{(0)})$ does encompass every possible isomorphism type of $\mathbf{u}_1^{(0)}\in\mathcal{R}_1(G)$ or $\mathbf{v}_1^{(0)}\in\mathcal{R}_1(H)$. Notice that $\mathbf{w}^{(0)}_1\in\mathcal{V}_{F^{(0)}_1(\tilde{G},\tilde{a})}^{i_1}$ is the isomorphism image (via $\varphi^{(0)}_1$) of some $\tilde{\mathbf{u}}^{(0)}_1\in\mathcal{R}_1(\tilde{G})$. When traversing over all $\tilde{G}$ and $\tilde{a}$, the value space of $\mathrm{atp}(\mathbf{w}_1^{(0)})$ is actually
    \begin{gather}
        \bigcup_{\tilde{G}\in\mathcal{G}}\{\mathrm{atp}(\tilde{\mathbf{u}}):\tilde{\mathbf{u}}\in\mathcal{R}_1(\tilde{G})\},
    \end{gather}
    which is definitely a superset of $\{\mathrm{atp}(\mathbf{u}_1^{(0)}):\mathbf{u}_1^{(0)}\in\mathcal{R}_1(G)\}$ and $\{\mathrm{atp}(\mathbf{v}_1^{(0)}):\mathbf{v}_1^{(0)}\in\mathcal{R}_1(H)\}$.

    Similarly, for $2\leqslant n\leqslant N$, the fact that $\forall F^{(0)}_{n}(\tilde{G},\tilde{a})\in\mathcal{S}_{n}^{(0)}$, $\mathrm{bIsoHom}(F^{(0)}_n(\tilde{G},\tilde{a}),G)=\mathrm{bIsoHom}(F^{(0)}_n(\tilde{G},\tilde{a}),H)$ implies that 
    \begin{gather}
        \big|\{\mathbf{u}_n^{(0)}\in\mathcal{V}_G^{i_n}:\mathrm{atp}(\mathbf{u}_n^{(0)})=\mathrm{atp}(\mathbf{w}_n^{(0)})\}\big|=\big|\{\mathbf{v}_n^{(0)}\in\mathcal{V}_H^{i_n}:\mathrm{atp}(\mathbf{v}_n^{(0)})=\mathrm{atp}(\mathbf{w}_n^{(0)})\}\big|.\label{atp_same_n0}
    \end{gather}
    With a completely analogous argument, we will find that there exists a bijection $f':\mathcal{R}_n(G)\rightarrow\mathcal{R}_n(H)$, such that for any $\mathbf{u}_n^{(0)}\in\mathcal{R}_n(G)$ and $\mathbf{v}_n^{(0)}=f'(\mathbf{u}_n^{(0)})\in\mathcal{R}_n(H)$, holds $\mathrm{atp}(\mathbf{u}_n^{(0)})=\mathrm{atp}(\mathbf{v}_n^{(0)})$. Assuming that Duplicator has not lost after the $(n-1)$-th putting move of the initialization round, the tuples $\mathbf{u}_{n-1}^{(0)}\in\mathcal{V}_G^{i_{n-1}}$ and $\mathbf{v}_{n-1}^{(0)}\in\mathcal{V}_H^{i_{n-1}}$ (occupied by pebbles $p_1,\ldots, p_{i_{n-1}}$ and $q_1,\ldots, q_{i_{n-1}}$ respectively) should have the same isomorphism type. If we restrict the bijection $f'$ on
    \begin{gather*}
        \{(\mathbf{u}_{n-1}^{(0)},\Delta\mathbf{u}_n^{(0)}):\Delta\mathbf{u}_n^{(0)}\in\mathcal{R}_n(G)\backslash\mathbf{u}_{n-1}^{(0)}\},
    \end{gather*}
    we can easily see that Duplicator can choose the bijection $f:\mathcal{R}_n(G)\backslash\mathbf{u}_{n-1}^{(0)}\rightarrow \mathcal{R}_n(H)\backslash\mathbf{v}_{n-1}^{(0)}$ such that $f'((\mathbf{u}_{n-1}^{(0)},\Delta \mathbf{u}_n^{(0)}))=(\mathbf{v}_{n-1}^{(0)},f(\Delta\mathbf{u}_n^{(0)}))$ for any $\Delta\mathbf{u}_n^{(0)}\in \mathcal{R}_n(G)\backslash\mathbf{u}_{n-1}^{(0)}$. By doing so, Duplicator will continue to survive the $n$-th putting move of the initialization round. By induction, we have proved that Duplicator can prevent herself from losing in the initialization round.

    The above technique can also be used to find Duplicator's strategy for the $m$-th putting move of the $T$-th update round, with $1\leqslant T\leqslant T^*$ and $1\leqslant m\leqslant M$. We only need to replace \eqref{atp_same_n0} with
    \begin{gather}
        \notag \big|\{(\mathbf{u}_N^{(T-1)},\mathbf{u}'^{(T)}_m)\in\mathcal{V}_G^{k+j_m}:\mathrm{atp}((\mathbf{u}_N^{(T-1)},\mathbf{u}'^{(T)}_m))=\mathrm{atp}(\mathbf{w}_m^{(T)})\}\big|\\
        =\big|\{(\mathbf{v}_N^{(T-1)},\mathbf{v}'^{(T)}_m)\in\mathcal{V}_H^{k+j_m}:\mathrm{atp}((\mathbf{v}_N^{(T-1)},\mathbf{v}'^{(T)}_m))=\mathrm{atp}(\mathbf{w}_m^{(T)})\}\big|,
    \end{gather}
    where $\mathrm{atp}(\mathbf{w}_m^{(T)})$ traverses over
    \begin{gather}
        \bigcup_{\tilde{G}\in\mathcal{G}}\{\mathrm{atp}((\tilde{\mathbf{u}},\tilde{\mathbf{u}}')):\tilde{\mathbf{u}}\in\mathcal{R}(\tilde{G}),\tilde{\mathbf{u}}'\in\mathcal{F}_m(\tilde{G},\tilde{\mathbf{u}})\},
    \end{gather}
    as we consider all graphs $F^{(T)}_m(\tilde{G},\tilde{a})\in\mathcal{S}_m^{(T)}$. With an argument similar to above, one can show that Duplicator indeed survives the $m$-th putting move of the $T$-th update round, given that she has survived the $(m-1)$-th putting move. Finally, the removing move poses no threats to Duplicator, if she has not lost after all $M$ putting moves in an update round. 
    
    Combining all the above results, we conclude that Duplicator can ensure that she will not lose the game in the first $(T^*+1)$ rounds, a result contradicting the assumption that Spoiler can win no later than the $T^*$-th update round ends. Therefore, Lemma \ref{different_bisohom} is true.\\\emph{End of proof of Lemma \ref{different_bisohom}.}

    We now turn to proving \textbf{(a)} $\Rightarrow$ \textbf{(b)}. As in the above discussion, we assume that Spoiler wins the Ehrenfeucht-Fra\"iss\'e game on $G$ and $H$ no later than the $T^*$-th update round ends. Our desired result is that $\exists F\in\mathcal{S}$, $\mathrm{Hom}(F,G)\ne\mathrm{Hom}(F,H)$. In Lemma \ref{different_bisohom}, we have already proved something similar, i.e., $\exists F\in\mathcal{S}$, $\mathrm{bIsoHom}(F,G)\ne\mathrm{bIsoHom}(F,H)$. To bridge the gap, it suffices to find connections between $\mathrm{Hom}(F,G)$ and $\mathrm{bIsoHom}(F,G)$, for $F\in\mathcal{S}$ and $G\in\mathcal{G}$.

    \begin{definition}\label{bag_wise_homo}
        Assume that $F\in\mathcal{S}$ and $\mathbf{w}=(w_1,\ldots, w_s)\in\mathcal{V}_F^s$ is a bag in $F$. Further assume that $w_i\ne w_j$ and $\{w_i,w_j\}\notin \mathcal{E}_F$ for some $i,j\in[s]$. Define two operations on $F$:\\
        (i) \textbf{Bag extension.} Let $F'$ be the graph obtained by adding an edge $\{w_i,w_j\}$ to $F$. We say that $F'$ is obtained by \textbf{extending the bag} $\mathbf{w}$ in $F$.\\
        (ii) \textbf{Bag surjection image. } Let $F'$ be the graph obtained by removing node $w_j$ in $F$, and then adding an edge between $w_i$ and every node that is adjacent to $w_j$ but not adjacent to $w_i$ in $F$. We say that $F'$ is obtained by \textbf{taking surjection image of bag} $\mathbf{w}$ in $F$.
    \end{definition}

    From Definition \ref{bag_wise_homo}, one can easily see that if $F'$ is obtained by extending (or taking surjection image of) a bag $\mathbf{w}$ in $F\in\mathcal{S}$, then there exists an \emph{in-bag} homomorphism $h$ from $F$ to $F'$, which satisfies (i) $h$ is surjective, and (ii) $u\ne v\Leftrightarrow h(u)\ne h(v)$ and $\{u,v\}\in\mathcal{E}_F\Leftrightarrow \{h(u),h(v)\}\in\mathcal{E}_{F'}$ if no single bag in $F$ contains both $u$ and $v$. Furthermore, we have
    \begin{gather}
        \notag \{\mathrm{atp}(h(\tilde{\mathbf{w}}^{(\tilde{T})}_{\tilde{p}})):F\in\mathcal{S}\text{ and }F\text{ contains the bag }\tilde{\mathbf{w}}^{(\tilde{T})}_{\tilde{p}}\}\\
        \subseteq \{\mathrm{atp}(\tilde{\mathbf{w}}^{(\tilde{T})}_{\tilde{p}}):F\in\mathcal{S}\text{ and }F\text{ contains the bag }\tilde{\mathbf{w}}^{(\tilde{T})}_{\tilde{p}}\},\label{subset_relation_between_bags}
    \end{gather}
    by our assumption that $\mathcal{R}(\cdot)$ and $\mathcal{F}(\cdot, \cdot)$ are both closed under homomorphisms. Equation \eqref{subset_relation_between_bags} implies that the image of every bag in $F$ under $h$ has the isomorphism type of a valid bag. Therefore, we assert that $F'\in\mathcal{S}$. In other words, $\mathcal{S}$ is closed under bag extension and taking bag surjection images. 
    
    In the following, we use $\mathrm{bInHom}(F,F')$ to denote the number of in-bag homomorphisms from $F$ to $F'$, if $F=F'$, or if $F'$ can be obtained by
    taking a series of bag extension or bag surjection image operations on $F$, given $F,F'\in\mathcal{S}$. If neither of above is the case, we define $\mathrm{bInHom}(F,F')=0$. Notice that $\mathrm{bInHom}(F,F')\ne 0$ only if $F$ and $F'$ lie in the same set $\mathcal{S}^{(T)}_p$ for some $T$ and $p$.

    One can see that the following identity holds for any $F\in\mathcal{S}$ and $G\in\mathcal{G}$,
    \begin{gather}
        \mathrm{Hom}(F,G)=\sum_{F'\in\mathcal{S}}\frac{\mathrm{bInHom}(F,F')}{\mathrm{bAut}(F')}\cdot\mathrm{bIsoHom}(F',G),
    \end{gather}
    in which $\mathrm{bAut}(F')$ ($F'\in\mathcal{S}$) is the number of automorphisms of $F'$ that is an automorphism of $F'[\mathbf{w}]$ for any bag $\mathbf{w}$ in $F'$. The above identity can be rewritten in matrix form, namely
    \begin{gather}
        \mathrm{Hom}_G=\mathrm{bInHom}\cdot \mathrm{bAut}^{-1}\cdot \mathrm{bIsoHom}_G,
    \end{gather}
    where $\mathrm{Hom}_G=[\mathrm{Hom}(F,G)]_{F\in\mathcal{S}}$ and $\mathrm{bIsoHom}_G=[\mathrm{bIsoHom}(F,G)]_{F\in\mathcal{S}}$ are column vectors, given $G\in\mathcal{G}$, while $\mathrm{bInHom}=[\mathrm{bInHom}(F,F')]_{F,F'\in\mathcal{S}}$ and $\mathrm{bAut}=\mathrm{diag}[\mathrm{bAut}(F)]_{F\in\mathcal{S}}$ are $|\mathcal{S}|\times|\mathcal{S}|$ matrices. Since one can easily show that the binary relation $\preccurlyeq$ on $\mathcal{S}$ defined by $F\preccurlyeq F'\Leftrightarrow \mathrm{bInHom}(F,F')\ne 0$ is a partial order on $\mathcal{S}$, there is a proper ordering of graphs in $\mathcal{S}$ that makes $\mathrm{bInHom}$ an upper triangular matrix whose diagonal entries are all non-zero. Furthermore, every diagonal entry of $\mathrm{bAut}^{-1}$ is non-zero. We thus assert that $\mathrm{bInHom}\cdot \mathrm{bAut}^{-1}$ is a full-rank matrix. Hence, given $G,H\in\mathcal{G}$ with $\mathrm{bIsoHom}_G\ne\mathrm{bIsoHom}_H$, we have $\mathrm{Hom}_G\ne\mathrm{Hom}_H$. Since we already have $\mathrm{bIsoHom}_G\ne\mathrm{bIsoHom}_H$ from Lemma \ref{different_bisohom}, we finally arrive at the desired result that $\mathrm{Hom}_G\ne\mathrm{Hom}_H$, or that $\exists F\in\mathcal{S}$, $\mathrm{Hom}(F,G)\ne\mathrm{Hom}(F,H)$.

    \textbf{(b) $\bm{\Rightarrow}$ (a).} We prove by giving a winning strategy for Spoiler. Before our proof, we have to introduce some concepts associated with the Cops-Robber game.

    \begin{definition}\label{node_class}
        Let $F\in\mathcal{G}$ and $\mathbf{u}=(u_1,\ldots, u_k)\in\mathcal{V}_F^k$. Assume that $\{C_1,\ldots, C_K\}$ is the complete set of different CCEs of $F$ induced by $A=\{u_1,\ldots, u_k\}$. Namely, we require $C_i\cap C_j=\varnothing$ for any $1\leqslant i<j\leqslant K$, and $\bigcup_{i=1}^K C_i=\mathcal{E}_F$. \\
        (i) \textbf{Boundary nodes.} The set
        \begin{gather}
            \mathcal{B}_i=\{u\in A:\exists e_1\in C_i \text{ and }e_2\in C_j\text{ with }i\ne j,e_1\text{ and }e_2\text{ have }u\text{ as their common end}\}
        \end{gather}
        is called the set of \textbf{boundary nodes} of $C_i$. \\
        (ii) \textbf{Internal nodes.} The set
        \begin{gather}
            \mathcal{I}_i=\{u\in A:\exists e\in C_i, e\text{ has }u\text{ as one of its ends}\}-\mathcal{B}_i
        \end{gather}
        is called the set of \textbf{internal nodes} of $C_i$. 
    \end{definition}
    
    The following lemma is quite straightforward, but will be useful for our proof.

    \begin{lemma}\label{decompose_homo}
        Keep all assumptions in Definition \ref{node_class}. Further let $G\in\mathcal{G}$ and $\mathbf{v}=(v_1,\ldots, v_k)\in\mathcal{V}_G^k$. If $\mathcal{B}_i\cup\mathcal{I}_i=\{u_{c_{i,1}},\ldots, u_{c_{i,n_i}}\}$ for $i\in[K]$, with $1\leqslant c_{i,1}<\cdots<c_{i,n_i}\leqslant k$, we define the operator $\mathbf{b}_i$ such that $\mathbf{b}_i(\mathbf{u})=(u_{c_{i,1}},\ldots, u_{c_{i,n_i}})\in\mathcal{V}_F^{n_i}$ and $\mathbf{b}_i(\mathbf{v})=(v_{c_{i,1}},\ldots, v_{c_{i,n_i}})\in\mathcal{V}_G^{n_i}$. We have
        \begin{gather}
            \mathrm{Hom}(F^{\mathbf{u}}, G^{\mathbf{v}})=\prod_{i=1}^K \mathrm{Hom}(C_i^{\mathbf{b}_i(\mathbf{u})}, G^{\mathbf{b}_i(\mathbf{v})}).\label{prod_of_homo_num}
        \end{gather}
        With a slight abuse of notation, in equation \eqref{prod_of_homo_num} we use $C_i$ to mean the subgraph of $F$ edge-induced by edges in $C_i$.
    \end{lemma}

    In the following, we assume that Cops have a winning strategy for the Cops-Robber game on $F\in\mathcal{G}$. The case is the worst for Cops if no internal nodes become unoccupied in any unguarding move. Actually, if Cops have a winning strategy such that they first put a pebble $p$ on $u$, and $u$ remains internal until $u$ becomes unoccupied again in an unguarding move, then Cops can choose to put $p$ at an alternative boundary node (already occupied by a pebble) and still win. Therefore, we always assume that the aforementioned condition holds.

    Let $\mathrm{Hom}(F,G)\ne \mathrm{Hom}(F,H)$, for some $G,H\in\mathcal{G}$. We will state Spoiler's winning strategy for the Ehrenfeucht-Fra\"iss\'e game on $G$ and $H$. Briefly speaking, Cops will play the role of an \emph{oracle} that Spoiler should consult before taking each move. 

    Without loss of generality, we assume that $F$ is connected. Otherwise, the fact that Cops win the Cops-Robber game on $F$ implies that they must win on each connected component of $F$ separately. Moreover, $\mathrm{Hom}(F,G)\ne\mathrm{Hom}(F,H)$ implies that at least one connected component $F'$ of $F$ satisfies $\mathrm{Hom}(F',G)\ne\mathrm{Hom}(F',H)$. Then we can apply the proof on $F'$.
    
    Assume that for the first guarding move of the initialization round, Cops should put pebbles $p_1,\ldots, p_{i_1}$ at $\mathbf{u}_1=\Delta\mathbf{u}_1=(u_1,\ldots, u_{i_1})\in\mathcal{R}_1(F)$. Since $\mathcal{R}(\cdot)$ is closed under homomorphisms, every homomorphism from $F$ to $G$ (or $H$) should map $\mathbf{u}_1$ to some $\mathbf{v}_1\in\mathcal{R}_1(G)$ (or $\mathbf{w}_1\in\mathcal{R}_1(H)$). The fact that $\mathrm{Hom}(F,G)\ne \mathrm{Hom}(F,H)$ implies that if there exists a bijection $f:\mathcal{R}_1(G)\rightarrow \mathcal{R}_1(H)$, then we can find $\mathbf{v}_1\in\mathcal{R}_1(G)$ and $\mathbf{w}_1=f(\mathbf{v}_1)\in\mathcal{R}_1(H)$ such that
    \begin{gather}
        \mathrm{Hom}(F^{\mathbf{u}_1},G^{\mathbf{v}_1})\ne \mathrm{Hom}(F^{\mathbf{u}_1}, H^{\mathbf{w}_1}).\label{different_labeled_homo_1}
    \end{gather}
    Now let us consider Spoiler's strategy for the Ehrenfeucht-Fra\"iss\'e game on $G$ and $H$. Notice that Duplicator either fails to choose a bijection $f:\mathcal{R}_1(G)\rightarrow \mathcal{R}_1(H)$ for the first putting move of the initialization round, or has to choose $f$ such that equation \eqref{different_labeled_homo_1} holds. If the former is the case, Spoiler already wins. Otherwise, we let Spoiler put pebbles $p_1,\ldots, p_{i_1}$ at $\mathbf{v}_1\in\mathcal{R}_1(G)$ that makes \eqref{different_labeled_homo_1} hold, and Duplicator then puts pebbles $q_1,\ldots, q_{i_1}$ at $\mathbf{w}_1\in\mathcal{R}_1(H)$. By following the above strategy, Spoiler can maintain equation \eqref{different_labeled_homo_1} at the end of the first putting move of the initialization round, if he has not yet won at that time.

    To proceed, let us further assume that the set of nodes $\{u_1,\ldots, u_{i_1}\}$ splits $F$ into $K_1^{(0)}$ CCEs, denoted as $C_1,\ldots, C_{K_1^{(0)}}$. Using Lemma \ref{decompose_homo}, we can rewrite equation \eqref{different_labeled_homo_1} as 
    \begin{gather}
        \prod_{i=1}^{K_1^{(0)}}\mathrm{Hom}(C_i^{\mathbf{b}_i(\mathbf{u}_1)},G^{\mathbf{b}_i(\mathbf{v}_1)})\ne \prod_{i=1}^{K_1^{(0)}}\mathrm{Hom}(C_i^{\mathbf{b}_i(\mathbf{u}_1)},H^{\mathbf{b}_i(\mathbf{w}_1)}),
    \end{gather}
    which further implies $\exists i\in[K_1^{(0)}]$, such that
    \begin{gather}
        \mathrm{Hom}(C_i^{\mathbf{b}_i(\mathbf{u}_1)},G^{\mathbf{b}_i(\mathbf{v}_1)})\ne \mathrm{Hom}(C_i^{\mathbf{b}_i(\mathbf{u}_1)},H^{\mathbf{b}_i(\mathbf{w}_1)}).\label{different_cce_1}
    \end{gather}
    One can naturally see $C_i$ as the CCE chosen by Robber in the first guarding move of the initialization round, as he plays the Cops-Robber game. Now, there are two cases. For the first case, $C_i$ already contains only one edge whose both ends are occupied by pebbles, then equation \eqref{different_cce_1} directly implies that $\mathbf{v}_1$ and $\mathbf{w}_1$ have different isomorphism types. Therefore, Spoiler already wins. Otherwise, we are sure that whatever $C_i$ ($i\in[K_1^{(0)}]$) chosen by Robber, Cops have a way to respond properly, by choosing a corresponding $\Delta\mathbf{u}_2\in\mathcal{R}_2(F)\backslash\mathbf{u}_1$. Since $\mathcal{R}(\cdot)$ is closed under homomorphisms, any homomorphism from $F$ to $G$ (or $H$) should map $\mathbf{u}_2=(\mathbf{u}_1,\Delta\mathbf{u}_2)$ to some $\mathbf{v}_2\in\mathcal{R}_2(G)$ (or some $\mathbf{w}_2\in\mathcal{R}_2(H)$). Therefore, we can repeat the above procedure to give Spoiler's strategy for the second (and further the $n$-th, with $2\leqslant n\leqslant N$) putting move of the initialization round.
    
    Let $1\leqslant n\leqslant N$. Assume that after the $(n-1)$-th putting move of the initialization round, neither Duplicator (in the Ehrenfeucht-Fra\"iss\'e game) nor Robber (in the Cops-Robber game) has lost. Moreover, we assume that Spoiler can maintain
    \begin{gather}
        \mathrm{Hom}(C^{\mathbf{b}(\mathbf{u}_{n-1})},G^{\mathbf{b}(\mathbf{v}_{n-1})})\ne \mathrm{Hom}(C^{\mathbf{b}(\mathbf{u}_{n-1})}, H^{\mathbf{b}(\mathbf{w}_{n-1})}),\label{different_cce_2}
    \end{gather}
    where $\mathbf{u}_{n-1}$ is the $i_{n-1}$-tuple occupied by pebbles on $F$, and $C$ is the CCE held by Robber, both at the end of the $(n-1)$-th guarding move of the initialization round; $\mathbf{v}_{n-1}$ and $\mathbf{w}_{n-1}$ are the tuples occupied by pebbles $p_1,\ldots, p_{i_{n-1}}$ (on $G$) and $q_1,\ldots, q_{i_{n-1}}$ (on $H$), at the end of the $(n-1)$-th putting move of the initialization round; $\mathbf{b}$ is the operator corresponding to $C$, as defined in Lemma \ref{decompose_homo}. Notice that equation \eqref{different_cce_2} goes back to $\mathrm{Hom}(F,G)\ne\mathrm{Hom}(F,H)$ when $n=1$, or to \eqref{different_cce_1} when $n=2$. Using a completely analogous argument to the case of $n=1$, we can show that Spoiler either wins in the $n$-th putting move of the initialization round, or can maintain that
    \begin{gather}
        \mathrm{Hom}(C^{\mathbf{b}(\mathbf{u}_n)},G^{\mathbf{b}(\mathbf{v}_n)})\ne \mathrm{Hom}(C^{\mathbf{b}(\mathbf{u}_n)}, H^{\mathbf{b}(\mathbf{w}_n)}),
    \end{gather}
    after the $n$-th putting move, where the notations are similar to above except that we are now considering the $n$-th putting/guarding move. Similar arguments apply to the putting moves of the update rounds. 

    What we only need to take care of is the removing moves. Assume that after the $M$-th putting move of some update round, holds
    \begin{gather}
        \mathrm{Hom}(C^{\mathbf{b}((\mathbf{u}_N,\mathbf{u}'_M))},G^{\mathbf{b}((\mathbf{v}_N,\mathbf{v}'_M))})\ne \mathrm{Hom}(C^{\mathbf{b}((\mathbf{u}_N,\mathbf{u}'_M))}, H^{\mathbf{b}((\mathbf{w}_N,\mathbf{w}'_M))}).
    \end{gather}
    Now Spoiler must choose some $r_1^{(c)},\ldots, r_k^{(c)}$ with $c\in[C]$, such that only the tuples $(\mathbf{v}_N)_c(\mathbf{v}'_M)$ and $(\mathbf{w}_N)_c(\mathbf{w}'_M)$ will remain occupied on $G$ and $H$, respectively. The proof cannot move on to the next update round unless we have
    \begin{gather}
        \mathbf{b}((\mathbf{u}_N,\mathbf{u}'_M))=\mathbf{b}((\mathbf{u}_N)_c(\mathbf{u}'_M)).\label{unchanged_boundary}
    \end{gather}
    Namely, we require that the union of boundary and internal nodes of $C$ remains unchanged after the corresponding unguarding move. Since Cops can win the Cops-Robber game on $F$, Cops can keep $C$ itself unchanged after the unguarding move. Therefore, the boundary nodes of $C$ remain unchanged. Thanks to our assumption that no internal nodes become unoccupied after the unguarding move, we assert that the internal nodes of $C$ also remain unchanged. We thus see \eqref{unchanged_boundary} holds. So far, we have made the proof.
\end{proof}

\end{document}